\documentclass[sigconf,balance=false]{acmart}
\usepackage{amsmath}
\usepackage{booktabs}
\usepackage{graphicx}
\usepackage{array}
\usepackage{multirow}
\usepackage{tabularx}
\usepackage{algorithm}
\usepackage{algpseudocode}
\usepackage{dblfloatfix}
\usepackage{placeins}
\usepackage{balance}
\graphicspath{{figures/}{../figures/}}
\usepackage{enumitem}

\renewcommand{\dbltopfraction}{0.95}
\renewcommand{\dblfloatpagefraction}{0.80}

\newcommand{\bestresult}[1]{\textbf{#1}}
\newcommand{\secondresult}[1]{\underline{#1}}
\newcommand{\resulttableformat}{%
  \normalsize
  \setlength{\tabcolsep}{3pt}%
  \renewcommand{\arraystretch}{1.20}%
}

\AtBeginDocument{%
  }

\setcopyright{none}
\renewcommand\footnotetextcopyrightpermission[1]{}

\acmConference[KDD '27]{The 33rd ACM SIGKDD Conference on Knowledge
  Discovery and Data Mining}{August 1--5, 2027}{San Jose, CA, USA}
\acmBooktitle{Proceedings of the 33rd ACM SIGKDD Conference on Knowledge
  Discovery and Data Mining (KDD '27), AI for Sciences Track,
  August 1--5, 2027, San Jose, CA, USA}
\acmYear{2027}
\copyrightyear{2027}

\begin{document}

\title{Multi-Source Dynamic Graph Learning for Compound-Flood Forecasting in Managed Coastal Systems}

\author{Liangjun You}
\orcid{0009-0006-1427-5197}
\affiliation{%
  \institution{City University of Hong Kong (Dongguan)}
  \country{China}
}
\email{72540091@cityu-dg.edu.cn}

\author{Min Wu}
\affiliation{%
  \institution{A*STAR}
  \country{Singapore}
}
\email{wumin@a-star.edu.sg}

\author{Orlando Woods}
\affiliation{%
  \institution{Singapore Management University}
  \country{Singapore}
}
\email{orlandowoods@smu.edu.sg}

\author{Dongsheng Luo}
\orcid{0000-0003-4192-0826}
\affiliation{%
  \institution{Singapore Management University}
  \country{Singapore}
}
\email{luodongsheng01@gmail.com}

\begin{abstract}
Compound flooding in managed coastal systems is influenced by hydrological
conditions and water-management activity observed across multiple monitoring
stations. Current forecasting models can capture temporal dependencies with low
average errors, but global error metrics may conceal poor reproduction of
prolonged high-water plateaus that are relevant to flood early warning. Because
hydrometeorological and operational signals are distributed across heterogeneous
gages, single-site records do not fully represent high-water dynamics.
Nevertheless, unconstrained fusion of cross-site signals can degrade the
stability of local temporal forecasts. This work proposes an anchored
forecasting framework that incorporates cross-site information through state-
and lead-dependent bounded residual corrections. A multi-source regime
representation constructed from hydrometeorological and operational
observations adaptively calibrates inter-site relationships and correction
scales, enabling targeted cross-site adjustment while preserving the local
temporal forecast as a stable anchor. Beyond conventional global error
statistics, we evaluate event-scale high-water characteristics through the
temporal alignment of forecasted and observed high-water processes. Experiments
demonstrate that selectively integrating multi-station dynamic conditions
improves the prediction reliability of sustained high-water plateaus while
maintaining high accuracy during routine hydrological conditions, supporting
flood early warning and water-management decision support. The data and source
code used in this study are publicly available at
\url{https://github.com/YljyLjylJ125/Compound-Flood-Forecasting}.
\end{abstract}



\maketitle

\flushbottom
\setlength{\emergencystretch}{1em}

\section{Introduction}
\label{sec:introduction}

\begin{figure}[!t]
    \centering
    \IfFileExists{figures/multisource_event_process.png}{%
        \includegraphics[width=\columnwidth]
        {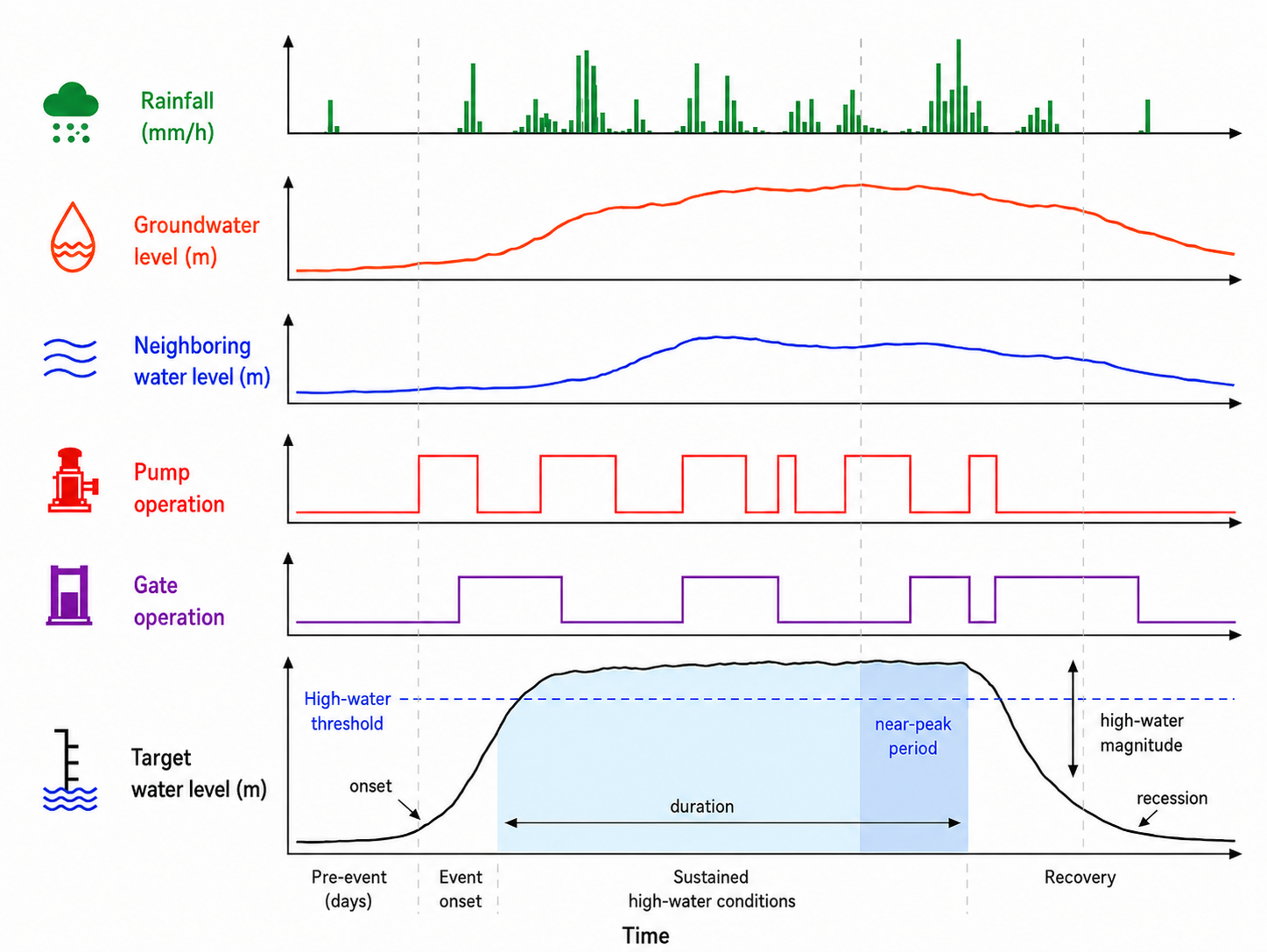}%
    }{%
        \fbox{\parbox[c][0.17\textheight][c]{0.92\columnwidth}{\centering
        Placeholder for \texttt{figures/multisource\_event\_process.png}}}%
    }
    \caption{Distributed observations across a high-water episode. The
    schematic follows environmental forcing, neighboring water levels, and
    control operations from onset through peak and recession.}
    \Description{Aligned conceptual time series for rainfall, groundwater,
    neighboring water levels, pump and gate operations, and the target water
    level during a sustained high-water episode.}
    \label{fig:event-process}
\end{figure}

\begin{figure*}[!t]
    \centering
    \IfFileExists{figures/managed_coastal_system.png}{%
        \includegraphics[width=0.92\textwidth]
        {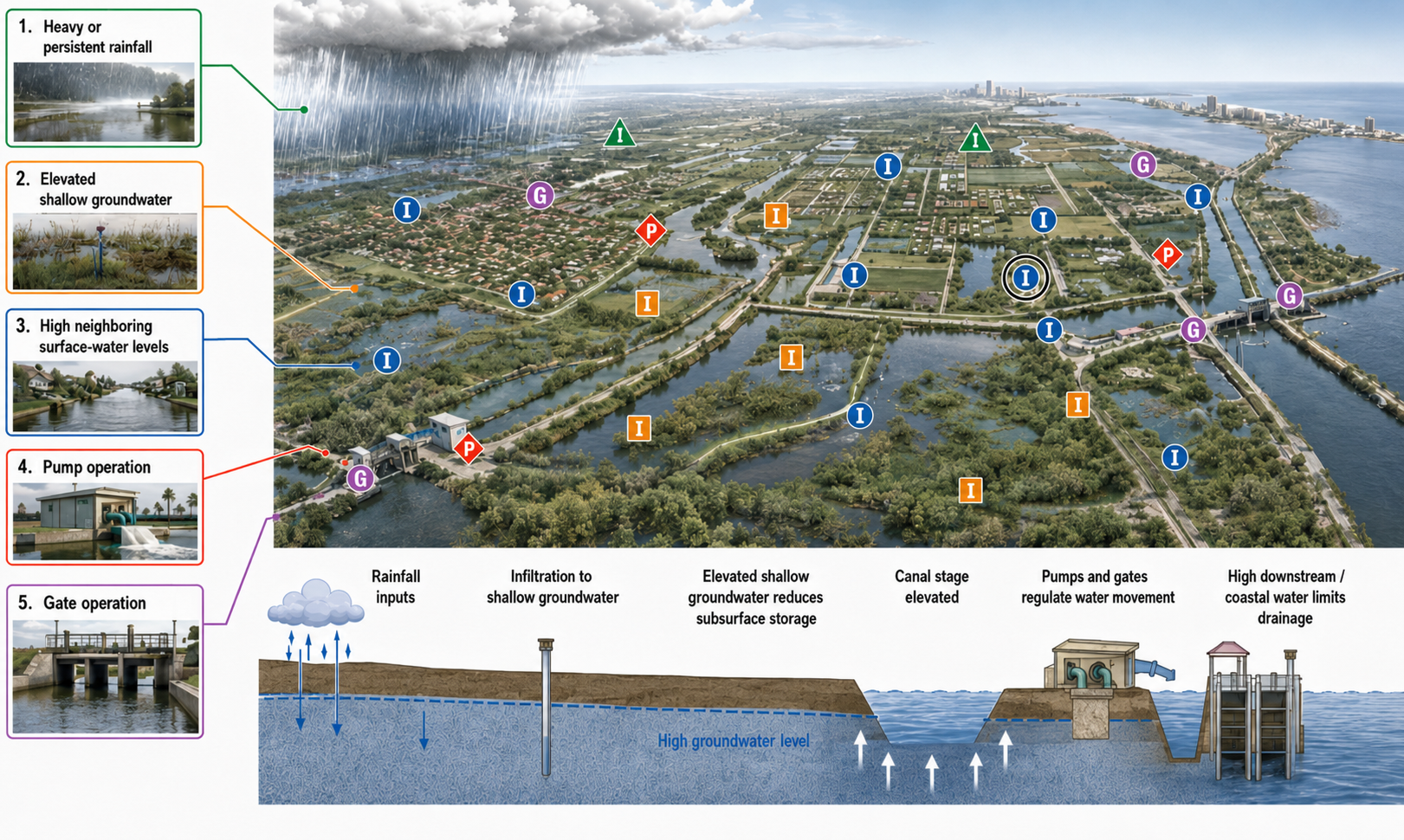}%
    }{%
        \fbox{\parbox[c][0.22\textheight][c]{0.92\textwidth}{\centering
        Placeholder for \texttt{figures/managed\_coastal\_system.png}}}%
    }
    \caption{Managed coastal monitoring and water-control system in South
    Florida, where rainfall, groundwater, surface water, pumps, and gates
    jointly shape water-level dynamics.}
    \Description{A South Florida landscape containing rainfall, shallow
    groundwater, canals, wetlands, a coastal boundary, pumps, gates, and
    distributed monitoring stations.}
    \label{fig:managed-system}
\end{figure*}

Flooding poses serious risks to human safety, infrastructure, and regional
economies~\cite{hirabayashi2013global}. Coastal flood forecasting becomes more
difficult when several drivers interact. Compound events may result from
simultaneous meteorological, hydrological, and coastal forcing
~\cite{wahl2015compound}. Their combined effects may exceed those of an
individual driver~\cite{zscheischler2020typology}. Heavy precipitation and
elevated coastal water levels are also expected to coincide more often under
climate change~\cite{bevacqua2020compound}. Forecasting water levels during
events therefore requires accurate temporal prediction and a principled
mechanism for deciding when observations elsewhere in the system should modify
a local forecast.

Water-control infrastructure also shapes flood behavior in managed coastal
basins. Rainfall supplies water, antecedent groundwater affects subsurface
storage, and high downstream or coastal levels can slow drainage. Pumps and
gates regulate flows through canals and storage areas
~\cite{obeysekera2011management}. These influences occur at different locations
and time scales: rainfall may coincide with elevated groundwater, neighboring
stages may remain high after rainfall ends, and control operations may alter
the rise or recession of local stage. Because their effects need not appear
simultaneously in the target record, forecasting becomes a distributed
observation problem in which different sites become informative at different
event stages. Figure~\ref{fig:event-process} summarizes this event-scale
sequence of distributed observations.

South Florida combines low relief, porous geology, heavy precipitation,
coastal influence, and extensive water management infrastructure
~\cite{jane2020southflorida}. Figure~\ref{fig:managed-system} summarizes this
setting. Surface-water stage (WATER) describes the evolving regional state;
rainfall (RAIN) records forcing; groundwater stage (WELL) represents antecedent
conditions; and pump (PUMP) and gate (GATE) records describe control activity.
The network is therefore a distributed observation of interacting environmental
and operational processes rather than a set of independent time series.

The prediction target throughout this paper is surface-water stage.
In managed coastal systems, locally sustained high stage can provide an early indication
of flood-relevant conditions. When it occurs alongside elevated coastal water or
hydrologic forcing, it is consistent with known compound-flood hazard
mechanisms~\cite{ward2018dependence}. We therefore treat high-water episodes as
a flood-relevant intermediate hydrologic state that provides a particularly hydrologically
meaningful reference for compound-flood monitoring and early situational
awareness.

The predictive value of these observations varies over time. Target stage is
strongly persistent in routine periods, whereas high-water episodes may depend
more on rainfall, groundwater, neighboring stages, and control activity. Errors
averaged over all station-hours are dominated by routine conditions and can
still easily obscure delayed onset, shortened duration, incorrect peak
magnitude, or different prediction delays despite similar pointwise
errors~\cite{jhin2024contime}. Evaluation should therefore consider both
full-record accuracy and high-water evolution.

We formulate cross-station information as a conditional departure from a local
temporal reference. A channel-wise PatchTST estimates the target
trajectory from its WATER history. An input-dependent heterogeneous graph then
summarizes the WATER, RAIN, WELL, PUMP, and GATE network. A multi-source regime
representation controls the correction by target and lead, while a
lead-dependent bound limits its magnitude. The temporal branch records local
support, and the graph branch records the additional departure supported by the
observed system state.

We compare the framework with temporal, multivariate, and graph forecasters over
separated periods, station groups, and forecast horizons using full-record and
episode-level measures. The main contributions are as follows:
\begin{itemize}[leftmargin=*,topsep=2pt,itemsep=1pt,parsep=0pt,partopsep=0pt]
    \item We formulate high-water forecasting in South Florida as a distributed
    observation problem that combines natural forcing, antecedent hydrologic
    state, neighboring water levels, and control activity while preserving a
    target-site temporal reference.
    \item We develop an anchored dynamic-graph forecaster that introduces
    network information through state- and lead-specific bounded residuals,
    making the local forecast and its network-supported correction separately
    inspectable.
    \item We evaluate whether these distributed observations improve both
    background water-level dynamics and sustained high-water processes.
    Training-derived station thresholds define episodes, and detection, onset,
    peak, and duration metrics complement MSE and MAE to assess flood-relevant
    process fidelity without using event labels during training.
\end{itemize}

\section{Related Work}

The forecasting problem connects two strands of prior work:
data-driven hydrological forecasting and spatiotemporal learning over
distributed monitoring networks.
Data-driven hydrological forecasting commonly uses meteorological inputs and
basin attributes to predict streamflow or discharge~\cite{shen2018review,
mosavi2018flood}. It has advanced rainfall--runoff, extreme-event, and
ungauged-basin prediction~\cite{kratzert2018rainfall,kratzert2019ungauged,
frame2022extreme,nearing2024global}, while related work has emphasized
uncertainty and scientific knowledge in learned environmental models
~\cite{klotz2022uncertainty,karpatne2017theory,read2019process,
willard2022scientific}. In managed coastal networks, surface-water stage,
rainfall, groundwater, neighboring stages, and control operations are observed
at heterogeneous locations. Their predictive value varies with system state,
target location, and lead time, favoring models that use distributed context
while retaining a clear local reference.

This setting differs from a single catchment with a prescribed drainage
hierarchy. A local stage record can be influenced by upstream, neighboring
water conditions, antecedent storage, and contemporaneous water-control
actions, while the available observations remain incomplete and unevenly
distributed. The forecasting task must therefore accommodate heterogeneous
evidence without assuming a static physical graph or a universal response rule.

Long-horizon forecasters use autoregressive, convolutional, basis-expansion,
and attention-based sequence models~\cite{salinas2020deepar,lai2018lstnet,
oreshkin2020nbeats,lim2021tft,zhou2021informer}. They extract temporal
structure well, but a single target sequence cannot represent the distributed
context of a managed water system. Early fusion can likewise obscure the
distinction between the target trajectory and external information.

This distinction is also important for scientific interpretation. A forecast
that directly combines all sources may be accurate without clarifying whether a
departure from recent local behavior is supported by the wider monitored state.
Keeping the local trajectory explicit makes the contribution of distributed
observations inspectable and limits the role of cross-station information to
revising, rather than replacing, the target-station forecast.

Graph forecasters propagate information across observed series using graph
signal processing and neural message passing~\cite{shuman2013graph,kipf2017gcn,
velickovic2018gat}. DCRNN, STGCN, ASTGCN, AGCRN, GMAN, and STSGCN model network
dependence from evolving observations~\cite{li2018dcrnn,yu2018stgcn,
guo2019astgcn,bai2020agcrn,zheng2020gman,song2020stsgcn}; hydrological studies
also use latent graphs and river topology~\cite{seman2026latentgnn,
kirschstein2024river}. These approaches generally do not distinguish the local
trajectory from its network-supported revision. This distinction is important
for heterogeneous monitoring networks with uneven coverage and time-varying
relations. Our formulation treats graph propagation as evidence for a bounded
departure from local temporal prediction.

\FloatBarrier
\section{Method}
\label{sec:method}

To resolve this tension, we decompose each forecast into a local WATER
trajectory and a bounded departure supported by the contemporaneous network
state. This formulation makes the information boundary between the local
reference and the network-supported correction explicit.

\subsection{Forecasting Objective and Local Reference}
\label{sec:problem}

Let $\mathcal{V}=\{1,\ldots,N\}$ be a heterogeneous monitoring network and
$\mathcal{V}_w\subset\mathcal{V}$ its $N_w$ WATER targets. For node $i$,
$c_i\in\mathcal{C}$ denotes its type, where $\mathcal{C}$ contains WATER,
RAIN, WELL, PUMP, and GATE, and $\mathbf{p}_i\in\mathbb{R}^2$ denotes its
coordinates.
Each series is standardized with its training-period mean and standard deviation.
$\mathbf{X}_t$ and $\mathbf{Y}_t$ below denote standardized values. At issue
time $t$, $\mathbf{X}_t\in\mathbb{R}^{N\times L}$ contains the $L$-hour
histories of all nodes. Together with its observation mask
$\mathbf{M}_t\in\{0,1\}^{N\times L}$, node types $\mathbf{c}$, and standardized
coordinates $\bar{\mathbf{P}}$, the model predicts the next $H$ hourly WATER levels
$\widehat{\mathbf{Y}}_t\in\mathbb{R}^{N_w\times H}$.
Predictions are transformed back to the original station units before error and
event evaluation. No observation after $t$ is used in either normalization or
forecasting.

Directly mapping all observations to the target would still leave local
persistence and cross-station information tightly entangled. Instead, we write
the forecast as a local trajectory plus a network-supported departure:
\begingroup
\setlength{\abovedisplayskip}{2pt plus 1pt minus 1pt}
\setlength{\belowdisplayskip}{2pt plus 1pt minus 1pt}
\setlength{\abovedisplayshortskip}{1pt plus 1pt}
\setlength{\belowdisplayshortskip}{2pt plus 1pt minus 1pt}
\begin{equation}
    \widehat{\mathbf{Y}}_t
    =\mathbf{A}_t+\mathbf{C}_t.
\label{eq:decomposition}
\end{equation}
\endgroup
Here $\mathbf{A}_t$ is computed independently for each target WATER station,
whereas $\mathbf{C}_t$ may use all heterogeneous nodes. The first term records
what can be predicted from local persistence alone. The second represents a
possible departure supported by the wider monitoring system. Both its activation
and its admissible magnitude are determined from observations available at issue
time, without future event labels.

For target WATER node $i$, the channel-wise PatchTST centers the local history
at its latest observed value. It then encodes overlapping patches with
parameters shared across target stations and restores the latest value after
decoding. Let $\mathbf{x}^{w}_{t,i}\in\mathbb{R}^{L}$ denote the local WATER
history ending at $t$. The anchor forecast is
\begin{equation}
    \mathbf{A}_{t,i,:}
    =
    X^{w}_{i,t}\mathbf{1}_{H}
    +
    \mathcal{D}_{A}\!\left(
        \mathcal{E}_{A}\!\left(
            \operatorname{Unfold}_{p,s}\!\left(
                \mathbf{x}^{w}_{t,i}
                -
                X^{w}_{i,t}\mathbf{1}_{L}
            \right)
        \right)
    \right).
\label{eq:anchor}
\end{equation}
The anchor has no access to other stations, node types, coordinates, or
non-WATER variables. Keeping this information boundary explicit makes the graph
contribution identifiable: any change relative to $\mathbf{A}_t$ must originate
from context unavailable to the local forecaster. The anchor defines the
locally supported trajectory that the network branch must preserve or revise.

\begin{table*}[t]
    \caption{Full-record MAE and MSE across forecast horizons of 1, 3, 5, and
    7 days. MAE preserves the linear contribution of each absolute stage
    deviation, whereas MSE squares the deviations and consequently gives
    greater influence to occasional large forecast errors.}
    \label{tab:full-record-accuracy}
    \centering
    \resulttableformat
    \resizebox{\textwidth}{\height}{%
    \begin{tabular}{@{}lcccccccc@{}}
        \toprule
        & \multicolumn{4}{c}{\textbf{MAE ($\times 10^{-2}$)}}
        & \multicolumn{4}{c}{\textbf{MSE ($\times 10^{-2}$)}} \\
        \cmidrule(lr){2-5}\cmidrule(lr){6-9}
        Model & 1D & 3D & 5D & 7D & 1D & 3D & 5D & 7D \\
        \midrule
        Ours
        & \bestresult{5.004$\pm$0.033}
        & \bestresult{9.031$\pm$0.098}
        & \bestresult{11.719$\pm$0.128}
        & \bestresult{14.135$\pm$0.137}
        & \bestresult{1.656$\pm$0.019}
        & \bestresult{4.043$\pm$0.040}
        & \bestresult{5.975$\pm$0.081}
        & \bestresult{7.680$\pm$0.086} \\
        NLinear
        & 5.609$\pm$0.053
        & 9.828$\pm$0.090
        & 12.854$\pm$0.146
        & 15.230$\pm$0.173
        & 1.902$\pm$0.017
        & 4.469$\pm$0.045
        & 6.641$\pm$0.077
        & 8.527$\pm$0.097 \\
        PatchTST
        & \secondresult{5.073$\pm$0.038}
        & \secondresult{9.324$\pm$0.126}
        & \secondresult{12.324$\pm$0.160}
        & \secondresult{14.743$\pm$0.142}
        & \secondresult{1.709$\pm$0.019}
        & \secondresult{4.248$\pm$0.049}
        & \secondresult{6.381$\pm$0.087}
        & \secondresult{8.282$\pm$0.107} \\
        iTransformer
        & 6.005$\pm$0.051
        & 10.210$\pm$0.077
        & 13.240$\pm$0.137
        & 15.664$\pm$0.132
        & 2.120$\pm$0.029
        & 4.723$\pm$0.074
        & 6.898$\pm$0.096
        & 8.806$\pm$0.113 \\
        TimesNet
        & 7.012$\pm$0.093
        & 10.900$\pm$0.166
        & 13.658$\pm$0.175
        & 16.026$\pm$0.205
        & 2.434$\pm$0.032
        & 5.037$\pm$0.066
        & 7.149$\pm$0.113
        & 9.244$\pm$0.164 \\
        FourierGNN
        & 6.390$\pm$0.054
        & 10.970$\pm$0.095
        & 14.017$\pm$0.166
        & 16.784$\pm$0.187
        & 1.959$\pm$0.018
        & 4.783$\pm$0.051
        & 7.005$\pm$0.088
        & 9.466$\pm$0.140 \\
        MTGNN
        & 8.587$\pm$0.118
        & 13.162$\pm$0.169
        & 17.256$\pm$0.273
        & 19.949$\pm$0.279
        & 6.790$\pm$0.099
        & 14.067$\pm$0.193
        & 18.279$\pm$0.224
        & 21.266$\pm$0.230 \\
        AutoTimes
        & 6.037$\pm$0.051
        & 10.087$\pm$0.096
        & 12.990$\pm$0.176
        & 15.324$\pm$0.208
        & 2.046$\pm$0.024
        & 4.599$\pm$0.066
        & 6.710$\pm$0.108
        & 8.579$\pm$0.147 \\
        Graph WaveNet
        & 6.181$\pm$0.076
        & 11.492$\pm$0.167
        & 14.611$\pm$0.235
        & 17.756$\pm$0.267
        & 3.072$\pm$0.035
        & 8.728$\pm$0.115
        & 14.280$\pm$0.195
        & 19.864$\pm$0.271 \\
        \bottomrule
    \end{tabular}%
    }
\end{table*}

\begin{table*}[t]
    \caption{Episode F1 and Onset MAE across forecast horizons of 1, 3, 5, and
    7 days. Episode F1 summarizes detection across matched, missed, and falsely
    predicted episodes, while Onset MAE measures start-time error among matched
    episodes. Station-specific training-period thresholds use $q=0.95$.}
    \label{tab:event-detection-onset}
    \centering
    \resulttableformat
    \resizebox{\textwidth}{\height}{%
    \begin{tabular}{@{}lcccccccc@{}}
        \toprule
        & \multicolumn{4}{c}{\textbf{Episode F1}}
        & \multicolumn{4}{c}{\textbf{Onset MAE (h)}} \\
        \cmidrule(lr){2-5}\cmidrule(lr){6-9}
        Model & 1D & 3D & 5D & 7D & 1D & 3D & 5D & 7D \\
        \midrule
        Ours
        & \bestresult{0.833$\pm$0.006}
        & \bestresult{0.652$\pm$0.007}
        & \bestresult{0.543$\pm$0.006}
        & \bestresult{0.470$\pm$0.004}
        & \bestresult{0.350$\pm$0.005}
        & \bestresult{1.658$\pm$0.021}
        & \bestresult{2.318$\pm$0.023}
        & \bestresult{4.121$\pm$0.036} \\
        NLinear
        & 0.806$\pm$0.009
        & 0.629$\pm$0.007
        & 0.519$\pm$0.007
        & 0.450$\pm$0.005
        & 0.459$\pm$0.006
        & 2.155$\pm$0.030
        & 4.135$\pm$0.057
        & 6.683$\pm$0.095 \\
        PatchTST
        & \secondresult{0.820$\pm$0.011}
        & \secondresult{0.639$\pm$0.009}
        & \secondresult{0.528$\pm$0.007}
        & \secondresult{0.457$\pm$0.005}
        & \secondresult{0.363$\pm$0.004}
        & \secondresult{1.730$\pm$0.016}
        & 3.227$\pm$0.031
        & 5.128$\pm$0.050 \\
        iTransformer
        & 0.786$\pm$0.007
        & 0.604$\pm$0.005
        & 0.509$\pm$0.005
        & 0.439$\pm$0.004
        & 0.499$\pm$0.006
        & 2.145$\pm$0.023
        & 5.131$\pm$0.051
        & 8.481$\pm$0.113 \\
        TimesNet
        & 0.757$\pm$0.010
        & 0.590$\pm$0.006
        & 0.503$\pm$0.007
        & 0.441$\pm$0.006
        & 0.800$\pm$0.013
        & 2.844$\pm$0.042
        & 6.095$\pm$0.095
        & 9.812$\pm$0.184 \\
        FourierGNN
        & 0.772$\pm$0.008
        & 0.573$\pm$0.006
        & 0.481$\pm$0.004
        & 0.385$\pm$0.004
        & 0.587$\pm$0.007
        & 1.873$\pm$0.027
        & \secondresult{2.484$\pm$0.032}
        & \secondresult{4.386$\pm$0.058} \\
        MTGNN
        & 0.717$\pm$0.011
        & 0.548$\pm$0.007
        & 0.422$\pm$0.006
        & 0.363$\pm$0.005
        & 1.203$\pm$0.018
        & 2.187$\pm$0.040
        & 4.088$\pm$0.077
        & 5.463$\pm$0.102 \\
        AutoTimes
        & 0.795$\pm$0.007
        & 0.619$\pm$0.008
        & 0.517$\pm$0.006
        & 0.440$\pm$0.005
        & 0.949$\pm$0.012
        & 3.000$\pm$0.041
        & 5.902$\pm$0.091
        & 8.326$\pm$0.136 \\
        Graph WaveNet
        & 0.765$\pm$0.009
        & 0.584$\pm$0.006
        & 0.477$\pm$0.007
        & 0.411$\pm$0.004
        & 0.641$\pm$0.007
        & 2.420$\pm$0.039
        & 4.872$\pm$0.052
        & 7.536$\pm$0.128 \\
        \bottomrule
    \end{tabular}%
    }
\end{table*}

\begin{table*}[t]
    \caption{Peak Magnitude MAE and Duration MAE across forecast horizons of 1,
    3, 5, and 7 days. Peak Magnitude MAE measures peak-stage error, while
    Duration MAE measures episode-length error. Both are computed for matched
    episodes using station-specific training-period thresholds at $q=0.95$.}
    \label{tab:event-peak-duration}
    \centering
    \resulttableformat
    \resizebox{\textwidth}{\height}{%
    \begin{tabular}{@{}lcccccccc@{}}
        \toprule
        & \multicolumn{4}{c}{\textbf{Peak Magnitude MAE}}
        & \multicolumn{4}{c}{\textbf{Duration MAE (h)}} \\
        \cmidrule(lr){2-5}\cmidrule(lr){6-9}
        Model & 1D & 3D & 5D & 7D & 1D & 3D & 5D & 7D \\
        \midrule
        Ours
        & \bestresult{0.045$\pm$0.000}
        & \bestresult{0.066$\pm$0.001}
        & \bestresult{0.081$\pm$0.001}
        & \bestresult{0.088$\pm$0.001}
        & \bestresult{2.297$\pm$0.028}
        & \bestresult{12.401$\pm$0.142}
        & \bestresult{23.873$\pm$0.237}
        & \bestresult{33.768$\pm$0.349} \\
        NLinear
        & 0.053$\pm$0.001
        & 0.080$\pm$0.001
        & 0.094$\pm$0.001
        & 0.101$\pm$0.001
        & 2.561$\pm$0.031
        & 13.988$\pm$0.178
        & 29.037$\pm$0.422
        & 45.871$\pm$0.594 \\
        PatchTST
        & \secondresult{0.046$\pm$0.000}
        & \secondresult{0.070$\pm$0.001}
        & \secondresult{0.084$\pm$0.001}
        & \secondresult{0.093$\pm$0.001}
        & \secondresult{2.354$\pm$0.031}
        & \secondresult{12.750$\pm$0.181}
        & 27.292$\pm$0.368
        & 43.101$\pm$0.635 \\
        iTransformer
        & 0.059$\pm$0.001
        & 0.086$\pm$0.001
        & 0.099$\pm$0.001
        & 0.110$\pm$0.001
        & 2.561$\pm$0.025
        & 13.856$\pm$0.161
        & 29.729$\pm$0.408
        & 48.431$\pm$0.676 \\
        TimesNet
        & 0.074$\pm$0.001
        & 0.104$\pm$0.001
        & 0.117$\pm$0.002
        & 0.129$\pm$0.002
        & 2.796$\pm$0.045
        & 13.790$\pm$0.247
        & 29.913$\pm$0.487
        & 48.410$\pm$0.846 \\
        FourierGNN
        & 0.066$\pm$0.001
        & 0.097$\pm$0.001
        & 0.109$\pm$0.001
        & 0.124$\pm$0.001
        & 2.623$\pm$0.031
        & 13.232$\pm$0.142
        & \secondresult{25.244$\pm$0.247}
        & \secondresult{35.824$\pm$0.343} \\
        MTGNN
        & 0.097$\pm$0.002
        & 0.110$\pm$0.002
        & 0.150$\pm$0.002
        & 0.141$\pm$0.002
        & 3.326$\pm$0.054
        & 15.383$\pm$0.273
        & 31.305$\pm$0.538
        & 39.548$\pm$0.618 \\
        AutoTimes
        & 0.057$\pm$0.001
        & 0.086$\pm$0.001
        & 0.098$\pm$0.002
        & 0.107$\pm$0.002
        & 2.536$\pm$0.035
        & 13.445$\pm$0.183
        & 28.250$\pm$0.421
        & 44.861$\pm$0.535 \\
        Graph WaveNet
        & 0.073$\pm$0.001
        & 0.103$\pm$0.001
        & 0.132$\pm$0.002
        & 0.132$\pm$0.001
        & 2.925$\pm$0.031
        & 14.580$\pm$0.229
        & 30.540$\pm$0.318
        & 47.260$\pm$0.744 \\
        \bottomrule
    \end{tabular}%
    }
\end{table*}

\begin{figure*}[t]
    \centering
    \includegraphics[width=\textwidth]
    {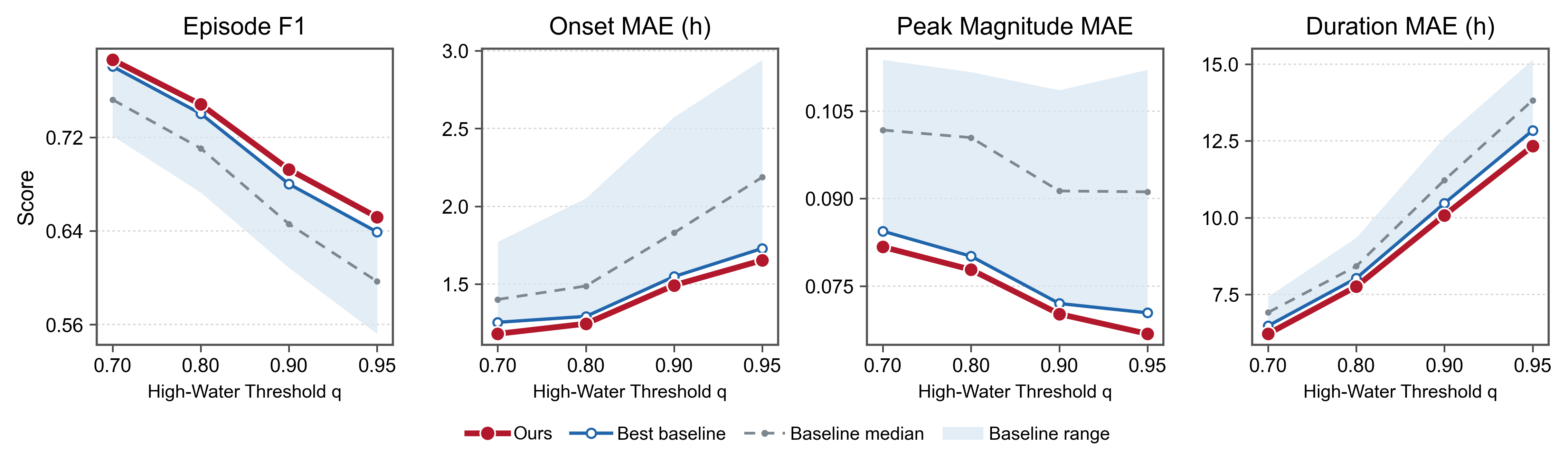}
    \caption{Sensitivity of episode-level performance to the station-specific
    high-water threshold for the $S_7$ split and 3-day forecast horizon.}
    \Description{Four line charts compare nine forecasting models at high-water
    quantiles 0.70, 0.80, 0.90, and 0.95. The proposed model has the highest
    Episode F1 and the lowest onset, peak-magnitude, and duration errors at all
    four thresholds.}
    \label{fig:threshold-sensitivity}
\end{figure*}

\begin{figure*}[t]
    \centering
    \includegraphics[width=\textwidth]{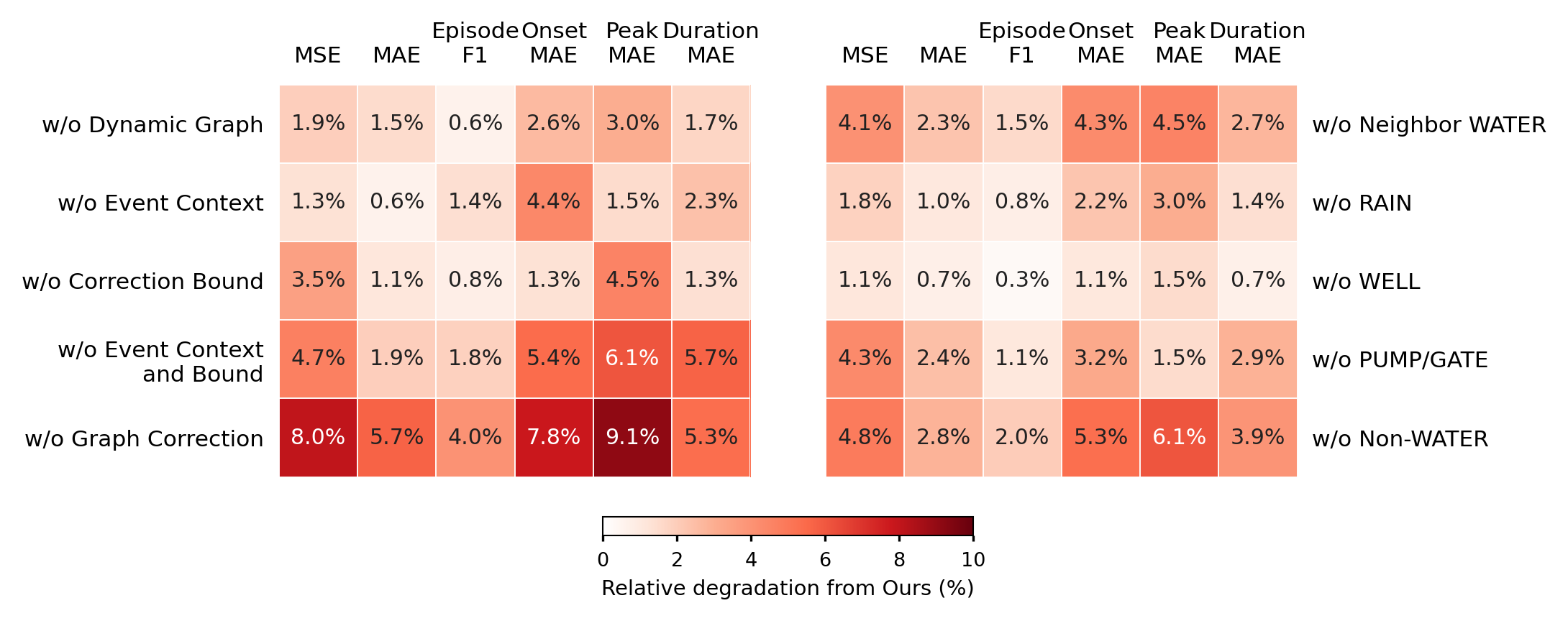}
    \caption{Relative performance degradation under architecture and
    observation-source ablations for the $S_7$ split, 3-day forecast horizon,
    and $q=0.95$ high-water threshold.}
    \Description{Two heatmaps show relative degradation in MSE, MAE, Episode
    F1, Onset MAE, Peak Magnitude MAE, and Duration MAE. The left panel removes
    dynamic graph construction, event context, the correction bound, their
    combination, or the complete graph-correction path. The right panel removes
    neighboring WATER, RAIN, WELL, PUMP and GATE, or all non-WATER observations.}
    \label{fig:ablation}
\end{figure*}

\subsection{State-Dependent Network Correction}

The correction branch begins with the context excluded from the
anchor. For node $i$ and historical step $\tau$, it concatenates the
observation, mask, type embedding, and standardized coordinate embedding:
\begin{equation}
    \mathbf{u}_{t,i,\tau}
    =[X_{i,\tau},M_{i,\tau},
    \mathbf{e}^{\mathrm{type}}_{c_i},\mathbf{e}^{\mathrm{loc}}_i].
\label{eq:node-input}
\end{equation}
A shared GRU then produces the window-dependent state
\begin{equation}
    \mathbf{h}_{t,i}
    =\operatorname{GRU}_{\phi}
    (\mathbf{u}_{t,i,t-L+1:t})_{\mathrm{last}}
    \in\mathbb{R}^{d}.
\label{eq:node-state}
\end{equation}
This state places observations with different temporal behavior in a common
representation while retaining their type and location. Because it is
recomputed for every input window, relations inferred from these states need
not remain strictly fixed across routine and high-water periods.

Useful sources can change with rainfall, groundwater, and control conditions.
We therefore score the directed relation from source node $j$ to receiving node
$i$ from their current states, while node type and distance restrict the search
through structural biases:
\begin{equation}
    s_{t,ij}
    =\frac{(\mathbf{W}_q\mathbf{h}_{t,i})^{\top}
    (\mathbf{W}_k\mathbf{h}_{t,j})}{\sqrt{d}}
    +B_{c_i,c_j}
    -\operatorname{softplus}(\rho)
    \|\bar{\mathbf{p}}_i-\bar{\mathbf{p}}_j\|_2.
\label{eq:graphscore}
\end{equation}
Let $\mathcal{N}^{K}_t(i)$ be the top-$K$ non-self sources under this score. The
row-normalized dynamic adjacency is
\begin{equation}
    A_{t,ij}
    =\frac{\mathbb{I}[j\in\mathcal{N}^{K}_t(i)]\exp(s_{t,ij})}
    {\sum_{k\in\mathcal{N}^{K}_t(i)}\exp(s_{t,ik})}.
\label{eq:dynamicadjacency}
\end{equation}
Because $s_{t,ij}$ depends directly on the current node states, both the retained
neighbors and their relative weights can change across forecast windows. The
type and coordinate terms act as structural biases when selecting and weighting
neighbors. They do not prescribe a fixed connectivity pattern. The graph
therefore acts as an input-dependent predictive structure rather than a fixed
description of the monitoring network.

The selected sources are aggregated with these dynamic weights, giving the
network-informed state
\begin{equation}
    \mathbf{z}_{t,i}=\mathbf{h}_{t,i}+
    \operatorname{MLP}_{g}\!\left(
    [\mathbf{h}_{t,i},
    \sum_j A_{t,ij}\mathbf{W}_v\mathbf{h}_{t,j}]\right).
\label{eq:graphupdate}
\end{equation}

The dynamic adjacency answers where network information is collected, but not
whether it should revise a particular target forecast. That decision depends on
the current hydrological and operational regime and may differ across targets
and forecast leads. We summarize the observed regime using two mask-normalized
statistics for each node type $c$: the mean absolute standardized magnitude over
the most recent $r=24$ hours, $\mu_{t,c}$, and the signed change over the full
lookback, $\delta_{t,c}$. Each statistic uses only observed nodes of type $c$ and
an $\epsilon>0$ denominator safeguard. No event labels or future thresholds are
used. An MLP maps the concatenated type-specific statistics to the multi-source
regime representation
$\mathbf{e}_t=\operatorname{MLP}_{e}
([\mu_{t,c},\delta_{t,c}]_{c\in\mathcal{C}})$.
For target $i\in\mathcal{V}_w$ and lead $\ell\in\{1,\ldots,H\}$, the regime gate is
\begin{equation}
    g_{t,i,\ell}
    =\sigma\!\left(
    \operatorname{MLP}_{\mathrm{gate}}
    ([\mathbf{h}_{t,i},\mathbf{e}_t])_{\ell}\right).
\label{eq:gate}
\end{equation}
The adjacency determines which sources contribute information. The gate maps
the current regime to the strength of correction for target $i$ at lead
$\ell$. The graph state can therefore be computed for every window without
forcing the same correction policy across routine and high-water conditions.

The gate modulates the correction but does not bound its displacement. A
decoder maps $\mathbf{z}_{t,i}$ to residual logits
$\mathbf{d}_{t,i}=\mathcal{D}_{R}(\mathbf{z}_{t,i})\in\mathbb{R}^{H}$. We use a
lead-dependent trust-region budget $\beta_{\ell}$ that increases linearly from
$\beta_{\min}$ at the first lead to $\beta_{\max}$ at the final lead. This
reflects increasing uncertainty with forecast distance while retaining a finite
departure at every lead. The bounded correction is
\begin{equation}
    C_{t,i,\ell}
    =g_{t,i,\ell}\beta_{\ell}\tanh(d_{t,i,\ell}).
\label{eq:correction}
\end{equation}
The three factors have distinct roles: $d_{t,i,\ell}$ supplies the direction and
unconstrained strength of the graph proposal, $g_{t,i,\ell}$ determines how much
the current regime supports intervention, and $\beta_{\ell}$ limits the largest
admissible departure from the local trajectory.
Since
$0<g_{t,i,\ell}<1$ and $|\tanh(\cdot)|\leq1$, the effective
regime-conditioned trust region satisfies
\begin{equation}
    |C_{t,i,\ell}|
    \leq g_{t,i,\ell}\beta_{\ell}
    \leq\beta_{\ell}.
\label{eq:bound}
\end{equation}
This inequality is the explicit trust-region guarantee: it limits departure
from the anchor without assuming that every graph correction is beneficial.
All parameters are optimized jointly using mean squared error over observed
WATER targets. Unavailable target entries are excluded by their observation
mask. Event labels and episode metrics
do not enter training, so improved high-water behavior arises from
how the model uses the observed network while controlling its departure from the
local reference. For batch size $B$, dense graph aggregation costs
$O(BN^2d)$.

\section{Experimental Design}
\label{sec:experiments}

\enlargethispage{2\baselineskip}

Experiments use SF2Bench~\cite{zheng2026compound}, an hourly
observation network from the managed coastal water system of South Florida.
The three chronological splits used in this study span 2010--2023 and, before
spatial partitioning, contain 585--716 WATER stations together with 870--964
RAIN, WELL, PUMP, and GATE stations. Each station is associated with an hourly
record and geographic coordinates, providing both temporal observations and
spatial context for the monitored system.
WATER stations provide forecast targets and describe the changing
 state of surface-water. RAIN represents meteorological forcing, WELL reflects
groundwater storage and drainage conditions, and PUMP and GATE record human
interventions that redistribute water through the canal network. The resulting
task is therefore a forecast of a coupled natural--human system rather than an
independent collection of time series.

Following the SF2Bench protocol, the main experiment combines the chronological $S_5$, $S_6$,
and $S_7$ splits, three station partitions within each split, and
forecast horizons of 1, 3, 5 and 7 days. The chronological splits cover
different historical periods, the station partitions change the monitored
locations, and the four horizons show how performance changes with lead time.
Forecasts are returned to the original WATER units before evaluation, allowing
full-record errors and episode characteristics to be interpreted by magnitude
and timing.

This design treats chronological period, spatial composition, and forecast
distance as complementary dimensions of generalization. The three splits expose
different combinations of hydrologic and operating conditions, while station
partitions change both the forecast targets and the network of observations
available to support them. Increasing lead time reduces the
support provided by the recent target history. Together, these settings
test whether distributed context remains useful beyond a particular historical
period, station arrangement, or short-term persistence regime.

At each forecast issue, every model receives the preceding 48 hours and
predicts the immediately following 1-, 3-, 5-, or 7-day trajectory.
Training-only normalization and explicit masks are used throughout input,
loss, and evaluation, so unavailable readings and observations after the issue
time are not treated as model inputs.

The proposed forecaster uses 64-dimensional hidden states, a top-20 dynamic
neighborhood, and dropout of 0.1. It is trained for 10 epochs with AdamW using a
batch size of 64, a learning rate of $10^{-3}$, and weight decay of $10^{-5}$;
the checkpoint with the lowest validation MSE is retained.

\subsection{Baselines}
\label{sec:baselines}

For comparison, we employ eight baseline models representing distinct
methodological paradigms:

\begin{itemize}[leftmargin=*]
    \item \textbf{NLinear}~\cite{zeng2023transformers} tests whether normalized
    persistence and trend explain future water-level evolution.
    \item \textbf{PatchTST}~\cite{nie2023patchtst} tests long-range local memory
    through channel-independent temporal patches.
    \item \textbf{iTransformer}~\cite{liu2024itransformer} learns dependence
    across observed series by treating each series as a token.
    \item \textbf{TimesNet}~\cite{wu2023timesnet} represents multi-period
    variation in recurring and nonstationary water-level dynamics.
    \item \textbf{FourierGNN}~\cite{yi2023fouriergnn} tests whether spectral
    structure captures cross-series coupling.
    \item \textbf{MTGNN}~\cite{wu2020mtgnn} learns inter-series connectivity
    and temporal propagation without prescribing a physical graph.
    \item \textbf{AutoTimes}~\cite{liu2024autotimes} provides an autoregressive
    comparison built on pretrained language-model representations.
    \item \textbf{Graph WaveNet}~\cite{wu2019graphwavenet} combines an adaptive
    graph with dilated temporal convolutions to model network transport and
    temporal memory jointly.
\end{itemize}

Together, these models cover local temporal forecasting, multivariate
dependence, spectral coupling, and learned graph propagation. Their comparison
tests whether a stable local reference benefits from a bounded network
correction when forecasting high-water processes.

\subsection{Evaluation Metrics}

Performance is evaluated at full-record and process scales. MSE emphasizes
large water-level deviations~\cite{gupta2009mse}, whereas MAE describes the
typical absolute error in the original WATER units.

High-water episodes are defined strictly relative to the local water-level
distribution of each target station. Let $x_{i,t}$ denote the original-unit
water level and $M_{i,t}$ its observation mask. For WATER station $i$, the
threshold is estimated exclusively from its valid training-period observations
as

\begin{equation}
    \tau_i^{(q)} = Q_q\!\left(
    \left\{x_{i,t}:t\in\mathcal{T}_{\mathrm{train}},\,M_{i,t}=1\right\}
    \right).
    \label{eq:episode-threshold}
\end{equation}
The main comparison uses each station's 95th percentile; $q=0.95$ therefore
marks the upper 5\% of its valid training distribution. A test hour is an
exceedance when the observed or predicted level
meets or exceeds its fixed training-derived threshold. An episode begins at the
first exceedance and contains at least three exceedance hours; runs separated by
no more than six hours are merged. Episodes are evaluated at 24-hour forecast-issue
intervals and temporally overlapping observed and predicted episodes are matched
one-to-one, prioritizing larger overlap.

Matching is one-to-one within each target-station forecast window. Episode F1
therefore reflects recovered, missed, and false-alarm processes, whereas the
three episode MAEs are calculated only for matched pairs.

The distinction between detection and matched-episode reconstruction is
essential for interpreting high-water forecasts. A small onset, peak, or
duration error alone could arise from a limited subset of episodes that were
recovered successfully, whereas Episode F1 also accounts for missed and false
episodes. Reporting the four measures together therefore asks whether a model
both identifies an elevated-water process and reproduces its subsequent
evolution. 

Episode F1 balances matched, missed, and false predicted episodes. For matched
episodes, Onset MAE measures start-time error in hours, Peak-Magnitude MAE the
peak-stage error in original units, and Duration MAE the episode-length error in
hours. Because episode frequency and persistence depend on the high-water
definition, we repeat the evaluation at $q\in\{0.70,0.80,0.90\}$ in addition to
$q=0.95$, using the same construction rules, $S_7$ split, and 3-day horizon.

Ablations use the same $S_7$, 3-day, and $q=0.95$ setting. Architecture variants
replace the dynamic graph with a fixed graph, remove event context, the
correction bound, both controls, or the complete graph correction. Source
variants retain the architecture but exclude neighboring WATER, RAIN, WELL,
PUMP/GATE, or all non-WATER observations.

\section{Results}
\label{sec:results}

The experiments address four linked scientific questions. First, does the
complete model improve full-record forecasts relative to strong local,
multivariate, and graph baselines? Second, does it better represent the
sustained high-water processes relevant to flood monitoring, including event
occurrence, onset, peak severity, and duration? Third, do these episode-level
findings persist when high water is defined at different positions in the
station-specific distribution? Fourth, which architectural mechanisms and
observation sources account for the improvement? We interpret the pointwise
and episode-level results together. Full-record error checks whether event
gains come at the cost of the background water-level trajectory, while episode
metrics describe the evolution of elevated-water conditions that can precede
or accompany flood impacts. Threshold sensitivity tests whether the comparison
depends on a single statistical definition of high water, and ablation analysis
separates the contributions of the correction design and the distributed
observation network.

\subsection{Overall Water-Level Forecast Accuracy}
\label{sec:results-overall}

Table~\ref{tab:full-record-accuracy} shows that the proposed model has the
lowest MSE and MAE at every horizon. At short lead times, the strong performance
of the local PatchTST baseline confirms that recent stage persistence explains
much of the immediate water-level evolution. As the horizon extends, however,
the wider separation from local forecasting indicates that rainfall,
groundwater, neighboring stage, and control activity provide increasing
predictive context when the recent target history becomes less informative.

The joint improvement in MAE and MSE is also scientifically relevant. It shows
that the model better represents both routine variations and occasional large
departures, rather than improving average behavior by smoothing the trajectory.
Because full-record errors are dominated by routine station-hours, these results
also indicate that network corrections do not destabilize background
water-level dynamics. This is consistent with the anchor preserving local
temporal support while the bounded graph branch revises only departures
supported by the observed system state.

\subsection{High-Water Episode Detection and Evolution}
\label{sec:results-episodes}

Table~\ref{tab:event-detection-onset} connects the forecast gains more directly
to high-water monitoring. The proposed model achieves the highest Episode F1
and the lowest Onset MAE throughout the evaluated horizons. These metrics
describe different aspects of the process: Episode F1 measures whether elevated
conditions are detected without excessive misses or false alarms, whereas Onset
MAE measures whether the transition begins at the correct time. Their joint
improvement shows that the model does not merely produce more threshold
crossings; it better locates the emergence of sustained high water.

This behavior is consistent with the distributed-observation hypothesis.
Rainfall, groundwater, neighboring WATER, and control activity can describe a
changing system state before its effects are fully expressed in the recent
history of the target station. The results do not assign causal influence to an
individual source, but they show that their combined state contains predictive
information about the occurrence and timing of high-water transitions.

Peak magnitude and duration complete this assessment. They describe the severity
and persistence of each matched episode.
Table~\ref{tab:event-peak-duration} examines the amplitude and persistence of
matched episodes. The proposed model has the lowest Peak Magnitude MAE and
Duration MAE across all horizons, showing that improved detection is accompanied
by a closer reconstruction of the episode itself. Peak magnitude describes the
maximum stage reached, while duration reflects how long storage, drainage, and
control conditions sustain elevated water. The duration task becomes more
difficult as uncertainty accumulates over longer horizons, yet the consistent
ranking indicates that distributed observations remain informative for the
plateau and recession of high-water processes.

These errors are conditional on matched episodes and should be read with
Episode F1. Their joint improvement indicates that the model better
represents episode occurrence, transition, amplitude, and persistence rather
than optimizing one event property at the expense of another. Together with the
full-record results, this supports a process-based view of forecast quality:
monitoring requires both accurate point values and a faithful description of how
elevated-water conditions evolve.

\subsection{Sensitivity to the High-Water Threshold}
\label{sec:results-threshold}

The operational meaning and frequency of a high-water episode depend on the
selected station-specific quantile. We therefore compare Episode F1, Onset MAE,
Peak Magnitude MAE, and Duration MAE under
$q\in\{0.70,0.80,0.90,0.95\}$ for the same $S_7$ split and 3D forecast horizon.
Lower quantiles describe more frequent elevated-water conditions, whereas
$q=0.95$ isolates the upper tail used in the main comparison.

Figure~\ref{fig:threshold-sensitivity} shows that the proposed model ranks first
on all four episode metrics at every threshold. Its advantage therefore extends
from relatively frequent elevated-water conditions to the less frequent upper
tail. The simultaneous ranking in detection, onset, peak, and duration further
indicates that the result is not produced by exchanging one aspect of process
fidelity for another.

This threshold stability is important because no single statistical quantile
fully represents the range of high-water conditions relevant to monitoring.
The results therefore place the main $q=0.95$ comparison within a broader
pattern and show that the predictive value of distributed observations is not
directly tied to one event definition. Absolute values across thresholds still
describe different episode populations and should not be interpreted as
repeated measurements of identical events.

\subsection{Ablation Study}
\label{sec:results-ablation}

\enlargethispage{\baselineskip}

To examine which representations of the monitored system support high-water
forecasting, we conduct two groups of ablations using $S_7$, the 3-day horizon,
and the main $q=0.95$ threshold. The architecture variants test three related
assumptions: cross-station dependence changes with the observed system state,
the need for correction depends on the current hydrologic and operational
context, and network-supported departures should remain bounded around the
local trajectory. Accordingly, \textit{w/o Dynamic Graph} uses fixed
cross-station relations; \textit{w/o Event Context} removes the regime
representation; \textit{w/o Correction Bound} removes the lead-dependent
bound; \textit{w/o Event Context and Bound} removes both controls; and
\textit{w/o Graph Correction} retains only the local anchor.
Figure~\ref{fig:ablation} reports degradation from the complete model, with
larger positive values indicating greater performance loss.

The full model performs best across all six metrics, while removing the graph
correction causes the largest degradation. Local stage persistence alone is
therefore insufficient to represent the evolution of high-water processes.
The loss under fixed cross-station relations indicates that the useful network
structure varies across hydrological and operational states. The remaining
variants further show that deciding when to revise the local trajectory and
limiting the magnitude of that revision are complementary requirements.

We next use leave-one-source-out analysis to examine the predictive information
carried by different parts of the observation network. The variants remove
neighboring WATER, RAIN, WELL, PUMP/GATE, or all non-WATER observations while
leaving the architecture unchanged. Every removal degrades performance.
Neighboring WATER describes the distributed surface-water state, RAIN records
external forcing, WELL reflects antecedent groundwater conditions, and
PUMP/GATE records managed redistribution. The larger loss after removing all
non-WATER observations indicates that these environmental and operational
measurements jointly add information beyond spatial WATER observations alone.

\section{Conclusion}

This study examined whether distributed observations in a managed coastal
system improve high-water forecasts beyond local stage persistence. The
anchored dynamic-graph forecaster retains an explicit target trajectory and
uses heterogeneous WATER, RAIN, WELL, PUMP, and GATE observations only through
state- and lead-dependent bounded corrections.

Across horizons, the model achieved lower full-record MSE and MAE, higher
Episode F1, and lower onset, peak, and duration errors than the baselines.
Ablations link these gains to dynamic cross-station relations, regime-aware
activation, bounded correction, and complementary observation sources. The
results show that the wider system state contains predictive information about
high-water evolution not fully expressed in recent target history, supporting
process-resolved monitoring and water-management planning.

Future work should test transferability across managed coastal systems and
rarer regimes, incorporate rainfall and coastal-boundary forecasts with planned
operations, and introduce hydraulic constraints and predictive uncertainty.

\section{Limitations and Ethical Considerations}
\label{sec:limitations-ethics}

The experiments cover one managed coastal system and a finite range of
observed environmental and operational conditions. The learned graph captures predictive
relations within this network, so its validity under other basins, sensor
configurations, and operating policies still requires external evaluation.
Station-specific training quantiles provide a reproducible high-water
definition, but not a regulatory flood stage, inundation extent, or damage
measure; they nevertheless characterize sustained high stage relevant to
compound-flood preparedness.

The dataset contains no personal or human-subject records, but pump and gate
data may potentially reveal sensitive critical-infrastructure information and should follow provider
terms. Sensor coverage, missing observations, and historical operations can
introduce location- or regime-specific bias. Operational use should include
explicit local threshold checks, calibrated uncertainty, and routine hydrologist and water
manager review; the model is decision support, not an autonomous controller.

\section{Generative AI Usage}

During the preparation of this work, the authors used ChatGPT and DeepL to
improve language clarity and readability. The authors verified the manuscript
and retain responsibility for its complete content.

\clearpage

\bibliographystyle{ACM-Reference-Format}
\bibliography{references}

\appendix

\setcounter{topnumber}{3}
\setcounter{bottomnumber}{2}
\setcounter{dbltopnumber}{2}
\setcounter{dblbotnumber}{1}
\renewcommand{\topfraction}{0.95}
\renewcommand{\bottomfraction}{0.90}
\renewcommand{\dbltopfraction}{0.95}
\renewcommand{\textfraction}{0.05}
\renewcommand{\floatpagefraction}{0.85}
\renewcommand{\dblfloatpagefraction}{0.85}

\newcommand{\appendixalgorithmcontent}{%
\begin{algorithm*}[!t]
    \caption{Training and Forecasting Procedure of the Anchored Dynamic-Graph Forecaster}
    \label{alg:anchored-forecaster}
    \small
    \begin{algorithmic}[1]
    \Require $\mathbf{X}_{t-L+1:t}$ and $\mathbf{M}_{t-L+1:t}$ available up to
    issue time $t$; node metadata $(\mathbf{c},\bar{\mathbf{P}})$; WATER targets
    $\mathcal{V}_w$; forecast horizon $H$
    \Ensure Multi-step WATER-level forecasts $\widehat{\mathbf{Y}}_t$
    \State Initialize $\Theta=\{f_A,f_\phi,g,f_r,f_c\}$ and the optimizer
    \For{each training forecast issue}
        \State $\mathbf{A}_t \gets f_A(\mathbf{X}^{w}_{t-L+1:t})$ \Comment{target WATER histories only}
        \State $\mathbf{h}_{t,i}\gets f_\phi(\mathbf{X}_{i,t-L+1:t},\mathbf{M}_{i,t-L+1:t},c_i,\bar{\mathbf{p}}_i),\quad i\in\mathcal{V}$
        \State $\mathbf{G}_t\gets g(\{\mathbf{h}_{t,i}\}_{i\in\mathcal{V}})$ \Comment{input-dependent heterogeneous graph}
        \State $\widetilde{\mathbf{h}}_{t,i}\gets\operatorname{Agg}(\mathbf{G}_t,\{\mathbf{h}_{t,j}\}_{j\in\mathcal{V}}),\quad i\in\mathcal{V}_w$
        \State $\mathbf{r}_t\gets f_r(\{\mathbf{h}_{t,i},\widetilde{\mathbf{h}}_{t,i}\}_{i\in\mathcal{V}_w})$ \Comment{current system regime}
        \State $(\boldsymbol{\alpha}_{t,i},\mathbf{b}_{t,i})\gets f_c(\mathbf{h}_{t,i},\widetilde{\mathbf{h}}_{t,i},\mathbf{r}_t),\quad i\in\mathcal{V}_w$
        \State $\mathbf{C}_{t,i}\gets\operatorname{clip}(\mathbf{b}_{t,i},-\boldsymbol{\alpha}_{t,i},\boldsymbol{\alpha}_{t,i})$ \Comment{bounded residual}
        \State $\widehat{\mathbf{Y}}_t\gets\mathbf{A}_t+\mathbf{C}_t$
        \State $\mathcal{L}\gets\operatorname{MaskedMSE}(\widehat{\mathbf{Y}}_t,\mathbf{Y}_t,\mathbf{M}^{w}_t)$ \Comment{no event labels or episode metrics}
        \State $\Theta\gets\operatorname{Update}(\Theta,\nabla_\Theta\mathcal{L})$
    \EndFor
    \State At each test issue $t$, evaluate the same forward map using only observations with time index $\tau\leq t$
    \State \Return $\widehat{\mathbf{Y}}_t$
    \end{algorithmic}
    \normalsize
\end{algorithm*}%
}

\section{Algorithmic Description}
\label{app:algorithm}
\appendixalgorithmcontent

The local anchor receives only the target station's WATER history. At the issue
time, the graph branch encodes all available heterogeneous observations, reconstructs a
state-dependent network, and produces a strictly bounded target- and lead-specific
residual correction. Event labels and episode metrics are used only after
forecasts are issued, for evaluation.

\section{Dataset Details}
\label{app:dataset}

SF2Bench represents the managed South Florida water system as an hourly,
heterogeneous observation network. Each record is
associated with a station type, geographic coordinates, an hourly value, and an
observation indicator. The archive contains 2,452 stations in five functional
groups. WATER stations measure surface-water stage and define the forecasting
targets; RAIN stations provide precipitation observations; WELL stations
describe groundwater level; and PUMP and GATE stations record water-control
activity. The original station values remain in their native units and are
standardized separately by series. A flood-observation repository distributed
with SF2Bench is not used for model fitting or for defining the high-water
episodes in this study.

For clarity, the station groups and their roles are summarized below rather
than in a separate inventory table:
\begin{itemize}[leftmargin=*,itemsep=1pt,topsep=2pt]
    \item \textbf{WATER (993 stations):} surface-water stage at canals,
    structures, and managed water bodies; forecast targets and neighboring
    surface-water context.
    \item \textbf{RAIN (349 stations):} hourly precipitation observations;
    external hydrologic forcing.
    \item \textbf{WELL (582 stations):} groundwater level; antecedent storage
    and drainage context.
    \item \textbf{PUMP (99 stations):} pump operating signals; managed water
    redistribution.
    \item \textbf{GATE (429 stations):} reported gate-control signals;
    hydraulic connectivity and control activity.
\end{itemize}

The archive is divided into chronological blocks whose station coverage grows
as the monitoring network expands. Table~\ref{tab:app-split-composition}
reports the pre-partition station inventory. The main experiments use the three
most recent blocks, $S_5$--$S_7$, because they provide the broadest joint
coverage of the five observation groups.

\begin{table}[!tbp]
    \caption{Station availability in the eight SF2Bench chronological blocks
    before applying the three spatial partitions.}
    \label{tab:app-split-composition}
    \centering
    \resulttableformat
    \setlength{\tabcolsep}{1.5pt}
    \begin{tabular*}{\columnwidth}{@{\extracolsep{\fill}}lrrrrrr@{}}
        \toprule
        Split & WATER & RAIN & WELL & PUMP & GATE & Total \\
        \midrule
        $S_0$ & 159 & 143 & 40  & 17 & 82  & 441 \\
        $S_1$ & 227 & 139 & 36  & 18 & 104 & 524 \\
        $S_2$ & 332 & 170 & 44  & 26 & 94  & 666 \\
        $S_3$ & 402 & 227 & 178 & 31 & 107 & 945 \\
        $S_4$ & 518 & 254 & 296 & 48 & 172 & 1288 \\
        $S_5$ & 585 & 216 & 333 & 65 & 256 & 1455 \\
        $S_6$ & 670 & 186 & 317 & 85 & 300 & 1558 \\
        $S_7$ & 716 & 194 & 352 & 89 & 329 & 1680 \\
        \bottomrule
    \end{tabular*}
\end{table}

Within each chronological block, SF2Bench supplies three geographically
distinct station subsets. They alter both the number of targets and the
composition of the observed environmental and operational context. The exact
inventories used by this paper are shown in
Table~\ref{tab:app-spatial-parts}. A zero PUMP count means that no pump station
falls inside that particular geographic subset; it does not indicate that pump
data were removed during preprocessing.

\begin{table}[!tbp]
    \caption{Composition of the three official spatial partitions used for
    $S_5$, $S_6$, and $S_7$.}
    \label{tab:app-spatial-parts}
    \centering
    \resulttableformat
    \setlength{\tabcolsep}{1pt}
    \begin{tabular*}{\columnwidth}{@{\extracolsep{\fill}}llrrrrrr@{}}
        \toprule
        Split & Part & WATER & RAIN & WELL & PUMP & GATE & Total \\
        \midrule
        $S_5$ & 0 & 20 & 12 & 71 & 0  & 7  & 110 \\
              & 1 & 72 & 31 & 40 & 11 & 28 & 182 \\
              & 2 & 10 & 6  & 19 & 0  & 3  & 38 \\
        \addlinespace[2pt]
        $S_6$ & 0 & 19 & 12 & 64 & 0  & 5  & 100 \\
              & 1 & 74 & 29 & 35 & 12 & 24 & 174 \\
              & 2 & 12 & 3  & 19 & 1  & 4  & 39 \\
        \addlinespace[2pt]
        $S_7$ & 0 & 22 & 12 & 71 & 0  & 7  & 112 \\
              & 1 & 80 & 30 & 24 & 13 & 24 & 171 \\
              & 2 & 30 & 9  & 53 & 3  & 17 & 112 \\
        \bottomrule
    \end{tabular*}
\end{table}

The temporal protocol is summarized in Table~\ref{tab:app-chronology}.
Windows are generated hourly with a 48-hour lookback and no gap between the
input and forecast intervals. The four forecast lengths are 24, 72, 120, and
168 hours; every valid WATER observation in the selected part is a target,
whereas RAIN, WELL, PUMP, and GATE observations enter as contextual inputs.

\begin{table}[!tbp]
    \caption{Chronological phases used in the main experiments.}
    \label{tab:app-chronology}
    \centering
    \resulttableformat
    \begin{tabular*}{\columnwidth}{@{\extracolsep{\fill}}llll@{}}
        \toprule
        Split & Train & Validation & Test \\
        \midrule
        $S_5$ & 2010--2012 & 2013 & 2014 \\
        $S_6$ & 2015--2017 & 2018 & 2019 \\
        $S_7$ & 2020--2021 & 2022 & 2023 \\
        \bottomrule
    \end{tabular*}
\end{table}

For each station, the mean and standard deviation are fitted from its training
phase and then reused for validation and test data. Standardized values are
clipped to $[-20,20]$ to limit the influence of isolated sensor artifacts.
Missing observations are retained through explicit masks; they are excluded
from the masked training loss and all reported errors. Predicted WATER values
are transformed back to their original station units before full-record and
episode evaluation. The station-specific episode threshold is also estimated
from valid training observations only.

\paragraph{Episode construction and matching.}
For each target station $i$, the high-water threshold is the $q$-quantile of
its valid training-period WATER values,
$\theta_i=Q_q({y_{i,t}:t\in\mathcal{T}_{\mathrm{train}}})$. The same fixed
threshold is applied to observed and forecast WATER levels after they are
returned to the original station units. Within each forecast window,
an exceedance begins when a valid value is at least $\theta_i$. Consecutive
exceedance runs separated by no more than six hourly steps are merged, and a
merged interval is retained only when it contains at least three valid
exceedance hours. Its onset is the first index of the merged interval, its
duration is the length of that interval, and its peak is the largest valid
WATER level within it. This construction permits a short interruption in an
sustained elevated-water process without treating it as two
unrelated episodes.

Observed and predicted episodes are extracted independently for every target
station at forecast issues sampled 24 hours apart. A true and predicted episode
are eligible to match only if their time intervals overlap. Candidate pairs are
ranked first by decreasing overlap and then by smaller onset separation; pairs
are accepted greedily so that each observed or predicted episode contributes to
at most one match. Episode F1 is computed from matched, missed, and false-alarm
episodes. Onset MAE, Peak Magnitude MAE, and Duration MAE are computed over
matched pairs only, respectively using the absolute difference in onset time,
peak WATER level, and interval length. Thus F1 reports whether elevated-water
processes are recovered, while the remaining measures describe the timing,
magnitude, and persistence of processes that are recovered by both series.

\paragraph{Masks, continuity, and forecast windows.}
Episode extraction never fills missing observations. A time step can enter an
initial exceedance run only when its observation mask is valid and its WATER
level reaches the station threshold. The six-hour merging rule is applied after
those initial runs are identified, so it can bridge a short interruption caused
by a below-threshold observation or an unavailable reading. The retained
episode must nevertheless contain the required three valid exceedance hours.
Its reported duration is the full span from the merged start to end, including
any short bridged interval; this makes duration error reflect the persistence
of the elevated-water process rather than only the count of isolated
exceedance samples. The peak is selected from valid observations in that span.
Stations with no valid training values have no well-defined threshold and are
excluded from episode evaluation rather than assigned a synthetic level.

Forecast windows are evaluated from issue times separated by 24 hours. For a
given issue, observed and forecast trajectories cover the same future horizon
and use the same valid-target mask. Overlapping windows from different issue
times are retained because they represent forecasts that would have been made
on different days with different available histories. The episode start and end
indices are therefore expressed relative to the corresponding forecast window;
the reported onset and duration errors are in hours. No observations after the
issue time enter the model input, the graph construction, the local anchor, or
the threshold estimate. Observations in the future horizon are used only after
the forecast is issued to form the evaluation target.

\paragraph{Aggregation and threshold sensitivity.}
For one model, split, station partition, and forecast horizon, all matched,
missed, and false-alarm records produced by the sampled issues and target
stations are first combined. Episode F1 is then calculated from the resulting
corresponding counts of true positives, false positives, and false negatives.
The three MAE measures average absolute errors over matched records, so they
quantify process reconstruction conditional on episode recovery; they are
interpreted alongside F1 rather than as substitutes for it. The reported
tabulated values in the main paper and Table~\ref{tab:app-threshold-sensitivity}
average the three official spatial partitions for the stated chronological
split and horizon.

The main comparison uses $q=0.95$, which emphasizes the upper tail of each
station's own training-period stage distribution. The threshold-sensitivity
analysis repeats the same construction at $q\in\{0.70,0.80,0.90,0.95\}$ for
$S_7$ and the 3-day horizon. Changing $q$ changes the frequency, typical
duration, and peak distribution of the evaluated episodes; the analysis is
therefore intended to assess whether the model ranking is retained under
different definitions of elevated water, rather than to interpret a numerical
change across thresholds as a single common error scale.

\paragraph{Recorded episode details.}
Each evaluation record retains the issue time, target-station identifier,
training-derived threshold, and match status. Matched records additionally
store signed onset, peak, and duration differences, whose absolute means yield
the reported MAEs. Forecasts from distinct 24-hour issue times remain separate
decisions, so overlapping future windows are not collapsed into one reconstructed
series. Thresholds, masks, and one-to-one matching are fixed within each
reported setting.

\section{Baseline and Training Configurations}
\label{app:configurations}

All models use the same training schedule, while model-specific settings retain
the conventions of their respective forecasting architectures. The common
optimization and episode-evaluation settings are listed in
Table~\ref{tab:app-common-training}; the instantiated architecture settings
are described below.

\newcommand{\appendixarchitecturesummary}{%
\begin{itemize}[leftmargin=1.3em,itemsep=1pt,topsep=2pt,parsep=0pt]
    \item \textbf{NLinear}: last-value-centered
    linear map from 48 inputs to $H$ outputs.
    \item \textbf{PatchTST}: channel-independent
    PatchTST with 3 encoder layers, hidden size 128, 16 heads, patch length 16,
    stride 8, RevIN, and dropout 0.2.
    \item \textbf{iTransformer}: hidden size 512,
    8 heads, 2 encoder layers, feed-forward size 2048, and dropout 0.1.
    \item \textbf{TimesNet}: two TimesBlocks, five
    dominant periods, model/feed-forward size 32, six inception kernels, and
    dropout 0.1.
    \item \textbf{FourierGNN}: spectral embedding 256,
    hidden size 512, three complex graph-convolution stages, and sparsity
    threshold 0.01.
    \item \textbf{MTGNN}: three layers, learned graph,
    node embedding 40, residual channels 32, skip channels 64, end channels 128,
    and dropout 0.3.
    \item \textbf{AutoTimes}: frozen GPT-2 base
    backbone; token length 24; two-layer encoder/decoder MLPs of width 256; and
    dropout 0.1.
    \item \textbf{Graph WaveNet}: adaptive adjacency with
    four blocks and four dilation layers; residual channels 32, skip channels
    256, end channels 512, and dropout 0.3.
\end{itemize}%
}

\setcounter{table}{6}

\begin{table}[!tbp]
    \caption{Common training and evaluation configuration.}
    \label{tab:app-common-training}
    \centering
    \resulttableformat
    \renewcommand{\arraystretch}{1.08}
    \begin{tabularx}{\columnwidth}{@{}>{\raggedright\arraybackslash}p{0.46\columnwidth}
        >{\raggedright\arraybackslash}X@{}}
        \toprule
        Setting & Value \\
        \midrule
        Lookback & 48 hourly steps \\
        Forecast horizons & 24, 72, 120, and 168 hourly steps \\
        Optimizer & AdamW \\
        Learning rate & $10^{-3}$ \\
        Weight decay & $10^{-5}$ \\
        Epochs & 10 \\
        Batch size & 64 \\
        Gradient-norm clip & 1.0 \\
        Random seeds & 3 \\
        Normalized clip & $[-20,20]$ \\
        Main episode quantile & $q=0.95$ \\
        Episode issue stride & 24 h \\
        Minimum episode duration & 3 h \\
        Episode merge gap & 6 h \\
        \bottomrule
    \end{tabularx}
\end{table}

\appendixarchitecturesummary

The eight baselines are exactly those used in the main comparison. Their
settings retain the defining temporal, spectral, or graph representation of
each architecture, while the common input window, forecast targets,
optimization schedule, and checkpoint rule remain fixed. NLinear and PatchTST
use direct temporal mappings and patch-based temporal representations,
respectively. iTransformer, TimesNet, FourierGNN, and AutoTimes provide
alternative multivariate or frequency-aware representations; MTGNN and Graph
WaveNet construct learned dependencies across the observed network.
\setcounter{table}{7}

\section{Detailed Results}
\label{app:detailed-results}

Tables~\ref{tab:app-s5-results}--\ref{tab:app-s7-results} report the
split-specific results underlying the aggregate comparisons in the main text.
Entries are means $\pm$ standard deviations over the three official spatial
partitions, so the three tables expose performance across chronological blocks,
network compositions, and forecast horizons.

Within a table, each metric is reported at the four lead times used in the
main study. MAE and MSE are calculated from the inverse-transformed WATER
forecasts, whereas the episode measures use the station-specific training
quantile $q=0.95$ and the matching procedure described in Appendix~A. The
tables therefore distinguish changes in ordinary stage error from changes in
episode occurrence, timing, peak level, and persistence.

The three chronological blocks provide complementary views of the same
forecasting task. Their station partitions change the monitored locations and
the mixture of auxiliary observations, while the lead times progressively
reduce the direct support available from the recent target history. For a fixed split and lead time, the model columns can be compared directly
within each metric block. The standard deviations summarize the variation over
the three official station partitions, not variation from random model
initialization. They consequently describe how a reported result changes with
the spatial composition of the monitored system, including the number and type
of contextual stations available to the forecast. The tables retain all six
metrics so that routine stage accuracy and high-water process reconstruction
can be inspected together for every chronological block.

\begin{table*}[!t]
    \caption{Detailed $S_5$ results averaged over the three spatial
    parts for the earliest evaluated historical block. Entries are means $\pm$
    standard deviations over Parts 0, 1, and 2. MAE and MSE use the
    $\times10^{-2}$ scale of the main text, and episode metrics use
    station-specific training thresholds at $q=0.95$.}
    \label{tab:app-s5-results}
    \centering
    \resulttableformat
    \setlength{\tabcolsep}{1pt}
    \renewcommand{\arraystretch}{1.70}
    \resizebox{\textwidth}{!}{%
    \begin{tabular}{@{}llrrrrrrrrr@{}}
        \toprule
        Metric & Lead & Ours & NLinear & PatchTST &
        \shortstack{iTrans-\\former} & TimesNet &
        \shortstack{Fourier\\GNN} & MTGNN & AutoTimes &
        \shortstack{Graph\\WaveNet} \\
        \midrule
        MAE $\times10^{-2}$ & 1D & \bestresult{5.174$\pm$0.032} & 5.738$\pm$0.049 & 5.194$\pm$0.041 & 6.023$\pm$0.046 & 7.017$\pm$0.096 & 6.369$\pm$0.049 & 6.237$\pm$0.088 & 6.214$\pm$0.051 & 5.683$\pm$0.068 \\
         & 3D & \bestresult{9.234$\pm$0.103} & 10.048$\pm$0.082 & 9.504$\pm$0.121 & 10.290$\pm$0.072 & 10.943$\pm$0.154 & 11.315$\pm$0.089 & 11.403$\pm$0.118 & 10.311$\pm$0.093 & 10.805$\pm$0.145 \\
         & 5D & \bestresult{11.895$\pm$0.118} & 13.021$\pm$0.139 & 12.498$\pm$0.162 & 13.414$\pm$0.129 & 13.728$\pm$0.167 & 14.281$\pm$0.149 & 14.593$\pm$0.205 & 13.270$\pm$0.163 & 14.321$\pm$0.229 \\
         & 7D & \bestresult{14.280$\pm$0.129} & 15.302$\pm$0.153 & 14.772$\pm$0.142 & 15.636$\pm$0.130 & 15.899$\pm$0.208 & 17.130$\pm$0.183 & 17.068$\pm$0.228 & 15.462$\pm$0.198 & 17.014$\pm$0.249 \\
        \midrule
        MSE $\times10^{-2}$ & 1D & \bestresult{1.904$\pm$0.023} & 2.199$\pm$0.018 & 1.958$\pm$0.020 & 2.400$\pm$0.033 & 2.661$\pm$0.035 & 2.104$\pm$0.020 & 1.994$\pm$0.028 & 2.308$\pm$0.027 & 1.975$\pm$0.022 \\
         & 3D & \bestresult{4.568$\pm$0.044} & 5.072$\pm$0.049 & 4.817$\pm$0.052 & 5.380$\pm$0.083 & 5.581$\pm$0.067 & 5.062$\pm$0.047 & 4.862$\pm$0.057 & 5.229$\pm$0.074 & 4.865$\pm$0.064 \\
         & 5D & \bestresult{6.694$\pm$0.084} & 7.409$\pm$0.076 & 7.164$\pm$0.088 & 7.689$\pm$0.101 & 7.919$\pm$0.115 & 7.293$\pm$0.080 & 7.516$\pm$0.079 & 7.595$\pm$0.117 & 7.163$\pm$0.090 \\
         & 7D & \bestresult{8.482$\pm$0.089} & 9.384$\pm$0.101 & 9.242$\pm$0.117 & 9.690$\pm$0.121 & 9.962$\pm$0.166 & 9.316$\pm$0.125 & 8.911$\pm$0.095 & 9.514$\pm$0.154 & 9.294$\pm$0.125 \\
        \midrule
        Episode F1 & 1D & \bestresult{0.867$\pm$0.007} & 0.835$\pm$0.009 & 0.858$\pm$0.011 & 0.825$\pm$0.007 & 0.827$\pm$0.011 & 0.815$\pm$0.007 & 0.780$\pm$0.012 & 0.827$\pm$0.007 & 0.801$\pm$0.009 \\
         & 3D & \bestresult{0.729$\pm$0.008} & 0.701$\pm$0.008 & 0.711$\pm$0.009 & 0.686$\pm$0.006 & 0.667$\pm$0.006 & 0.656$\pm$0.007 & 0.612$\pm$0.007 & 0.698$\pm$0.008 & 0.628$\pm$0.005 \\
         & 5D & \bestresult{0.630$\pm$0.006} & 0.602$\pm$0.007 & 0.621$\pm$0.008 & 0.603$\pm$0.006 & 0.606$\pm$0.008 & 0.556$\pm$0.004 & 0.508$\pm$0.007 & 0.609$\pm$0.007 & 0.560$\pm$0.008 \\
         & 7D & \bestresult{0.565$\pm$0.005} & 0.536$\pm$0.005 & 0.546$\pm$0.006 & 0.537$\pm$0.005 & 0.531$\pm$0.007 & 0.459$\pm$0.004 & 0.403$\pm$0.005 & 0.534$\pm$0.006 & 0.493$\pm$0.004 \\
        \midrule
        Onset MAE & 1D & \bestresult{0.367$\pm$0.005} & 0.433$\pm$0.005 & 0.432$\pm$0.005 & 0.493$\pm$0.006 & 0.705$\pm$0.012 & 0.419$\pm$0.005 & 0.397$\pm$0.005 & 0.874$\pm$0.011 & 0.598$\pm$0.006 \\
         & 3D & 1.491$\pm$0.019 & 1.842$\pm$0.026 & 1.523$\pm$0.013 & 1.830$\pm$0.019 & 2.343$\pm$0.030 & 1.633$\pm$0.025 & 2.359$\pm$0.040 & 2.881$\pm$0.036 & \bestresult{0.962$\pm$0.013} \\
         & 5D & \bestresult{2.150$\pm$0.018} & 3.820$\pm$0.051 & 3.322$\pm$0.029 & 4.015$\pm$0.037 & 4.850$\pm$0.075 & 2.340$\pm$0.028 & 4.262$\pm$0.073 & 4.827$\pm$0.072 & 4.480$\pm$0.044 \\
         & 7D & 3.431$\pm$0.030 & 6.035$\pm$0.084 & 5.459$\pm$0.050 & 5.486$\pm$0.066 & 6.852$\pm$0.119 & \bestresult{3.093$\pm$0.038} & 7.903$\pm$0.134 & 6.416$\pm$0.097 & 6.025$\pm$0.097 \\
        \midrule
        Peak MAE & 1D & \bestresult{0.058$\pm$0.000} & 0.071$\pm$0.001 & 0.060$\pm$0.000 & 0.074$\pm$0.001 & 0.096$\pm$0.001 & 0.079$\pm$0.001 & 0.087$\pm$0.001 & 0.072$\pm$0.001 & 0.092$\pm$0.001 \\
         & 3D & \bestresult{0.087$\pm$0.001} & 0.104$\pm$0.001 & 0.090$\pm$0.001 & 0.109$\pm$0.001 & 0.119$\pm$0.002 & 0.121$\pm$0.002 & 0.150$\pm$0.002 & 0.110$\pm$0.001 & 0.134$\pm$0.001 \\
         & 5D & \bestresult{0.110$\pm$0.001} & 0.120$\pm$0.001 & \bestresult{0.110$\pm$0.001} & 0.126$\pm$0.001 & 0.133$\pm$0.002 & 0.143$\pm$0.002 & 0.188$\pm$0.003 & 0.123$\pm$0.002 & 0.169$\pm$0.002 \\
         & 7D & \bestresult{0.116$\pm$0.002} & 0.130$\pm$0.001 & 0.122$\pm$0.001 & 0.131$\pm$0.001 & 0.142$\pm$0.002 & 0.173$\pm$0.002 & 0.178$\pm$0.003 & 0.136$\pm$0.002 & 0.169$\pm$0.001 \\
        \midrule
        Duration MAE & 1D & \bestresult{1.743$\pm$0.021} & 1.973$\pm$0.023 & 1.935$\pm$0.024 & 1.818$\pm$0.015 & 2.160$\pm$0.030 & 1.942$\pm$0.021 & 2.392$\pm$0.036 & 2.022$\pm$0.025 & 2.240$\pm$0.021 \\
         & 3D & 9.328$\pm$0.097 & 10.001$\pm$0.135 & \bestresult{8.667$\pm$0.114} & 9.027$\pm$0.108 & 9.960$\pm$0.186 & 11.129$\pm$0.109 & 10.259$\pm$0.176 & 9.581$\pm$0.123 & 11.312$\pm$0.167 \\
         & 5D & \bestresult{17.526$\pm$0.181} & 20.454$\pm$0.284 & 18.253$\pm$0.230 & 18.988$\pm$0.276 & 20.538$\pm$0.336 & 21.697$\pm$0.209 & 29.376$\pm$0.488 & 19.178$\pm$0.301 & 21.390$\pm$0.193 \\
         & 7D & \bestresult{25.159$\pm$0.277} & 32.558$\pm$0.379 & 30.599$\pm$0.458 & 30.487$\pm$0.436 & 33.886$\pm$0.623 & 36.493$\pm$0.311 & 39.814$\pm$0.574 & 29.570$\pm$0.315 & 33.492$\pm$0.471 \\
        \bottomrule
    \end{tabular}%
    }
\end{table*}

\begin{table*}[!t]
    \caption{Detailed $S_6$ results for the intermediate evaluated historical
    block. Entries are means $\pm$ standard deviations over the three spatial
    parts.}
    \label{tab:app-s6-results}
    \centering
    \resulttableformat
    \setlength{\tabcolsep}{1pt}
    \renewcommand{\arraystretch}{1.70}
    \resizebox{\textwidth}{!}{%
    \begin{tabular}{@{}llrrrrrrrrr@{}}
        \toprule
        Metric & Lead & Ours & NLinear & PatchTST &
        \shortstack{iTrans-\\former} & TimesNet &
        \shortstack{Fourier\\GNN} & MTGNN & AutoTimes &
        \shortstack{Graph\\WaveNet} \\
        \midrule
        MAE $\times10^{-2}$ & 1D & 5.001$\pm$0.033 & 5.388$\pm$0.052 & \bestresult{4.907$\pm$0.036} & 6.200$\pm$0.054 & 6.722$\pm$0.079 & 6.210$\pm$0.053 & 8.927$\pm$0.113 & 5.810$\pm$0.050 & 5.778$\pm$0.065 \\
         & 3D & \bestresult{8.854$\pm$0.085} & 9.610$\pm$0.090 & 9.116$\pm$0.118 & 10.241$\pm$0.085 & 10.855$\pm$0.158 & 10.589$\pm$0.097 & 13.293$\pm$0.178 & 9.846$\pm$0.102 & 12.358$\pm$0.178 \\
         & 5D & \bestresult{11.786$\pm$0.141} & 12.534$\pm$0.148 & 12.205$\pm$0.141 & 13.190$\pm$0.147 & 13.547$\pm$0.166 & 13.925$\pm$0.183 & 18.212$\pm$0.284 & 12.768$\pm$0.177 & 14.163$\pm$0.229 \\
         & 7D & \bestresult{14.258$\pm$0.134} & 14.949$\pm$0.176 & 14.625$\pm$0.140 & 15.665$\pm$0.124 & 16.046$\pm$0.210 & 16.645$\pm$0.183 & 20.062$\pm$0.297 & 15.192$\pm$0.223 & 17.282$\pm$0.260 \\
        \midrule
        MSE $\times10^{-2}$ & 1D & \bestresult{1.489$\pm$0.016} & 1.637$\pm$0.016 & 1.508$\pm$0.017 & 2.003$\pm$0.029 & 2.134$\pm$0.029 & 1.764$\pm$0.016 & 2.383$\pm$0.040 & 1.824$\pm$0.022 & 1.601$\pm$0.017 \\
         & 3D & \bestresult{3.641$\pm$0.035} & 3.975$\pm$0.040 & 3.832$\pm$0.042 & 4.352$\pm$0.063 & 4.679$\pm$0.064 & 4.074$\pm$0.045 & 4.867$\pm$0.060 & 4.140$\pm$0.057 & 4.779$\pm$0.064 \\
         & 5D & \bestresult{5.437$\pm$0.078} & 5.973$\pm$0.071 & 5.817$\pm$0.082 & 6.410$\pm$0.084 & 6.485$\pm$0.112 & 6.125$\pm$0.073 & 8.009$\pm$0.093 & 6.058$\pm$0.104 & 6.578$\pm$0.089 \\
         & 7D & \bestresult{7.053$\pm$0.085} & 7.735$\pm$0.097 & 7.540$\pm$0.094 & 8.206$\pm$0.100 & 8.667$\pm$0.170 & 7.782$\pm$0.109 & 9.699$\pm$0.091 & 7.846$\pm$0.142 & 8.117$\pm$0.122 \\
        \midrule
        Episode F1 & 1D & \bestresult{0.840$\pm$0.006} & 0.815$\pm$0.008 & 0.824$\pm$0.012 & 0.789$\pm$0.007 & 0.755$\pm$0.009 & 0.778$\pm$0.008 & 0.743$\pm$0.012 & 0.810$\pm$0.006 & 0.771$\pm$0.009 \\
         & 3D & \bestresult{0.641$\pm$0.006} & 0.614$\pm$0.007 & 0.628$\pm$0.010 & 0.577$\pm$0.005 & 0.582$\pm$0.006 & 0.544$\pm$0.006 & 0.562$\pm$0.006 & 0.598$\pm$0.008 & 0.572$\pm$0.006 \\
         & 5D & \bestresult{0.519$\pm$0.006} & 0.493$\pm$0.007 & 0.499$\pm$0.006 & 0.470$\pm$0.005 & 0.482$\pm$0.007 & 0.469$\pm$0.005 & 0.390$\pm$0.005 & 0.495$\pm$0.005 & 0.453$\pm$0.006 \\
         & 7D & \bestresult{0.433$\pm$0.004} & 0.418$\pm$0.005 & 0.424$\pm$0.005 & 0.393$\pm$0.004 & 0.414$\pm$0.005 & 0.379$\pm$0.004 & 0.404$\pm$0.005 & 0.416$\pm$0.004 & 0.384$\pm$0.004 \\
        \midrule
        Onset MAE & 1D & 0.288$\pm$0.004 & 0.386$\pm$0.005 & 0.267$\pm$0.003 & \bestresult{0.261$\pm$0.003} & 0.700$\pm$0.011 & 0.583$\pm$0.007 & 2.098$\pm$0.030 & 0.693$\pm$0.008 & 0.534$\pm$0.005 \\
         & 3D & 1.754$\pm$0.022 & 2.110$\pm$0.028 & 1.759$\pm$0.015 & 1.667$\pm$0.018 & 2.113$\pm$0.030 & \bestresult{0.807$\pm$0.012} & 2.093$\pm$0.041 & 2.582$\pm$0.036 & 2.119$\pm$0.035 \\
         & 5D & \bestresult{1.893$\pm$0.020} & 3.800$\pm$0.052 & 2.450$\pm$0.021 & 4.872$\pm$0.054 & 6.158$\pm$0.092 & 2.124$\pm$0.028 & 2.238$\pm$0.045 & 6.083$\pm$0.088 & 4.293$\pm$0.041 \\
         & 7D & 4.252$\pm$0.037 & 5.861$\pm$0.079 & 3.813$\pm$0.034 & 9.405$\pm$0.136 & 11.746$\pm$0.218 & \bestresult{3.707$\pm$0.048} & 5.490$\pm$0.113 & 8.703$\pm$0.147 & 7.654$\pm$0.139 \\
        \midrule
        Peak MAE & 1D & \bestresult{0.031$\pm$0.000} & 0.035$\pm$0.000 & 0.032$\pm$0.000 & 0.037$\pm$0.000 & 0.041$\pm$0.001 & 0.041$\pm$0.000 & 0.062$\pm$0.001 & 0.037$\pm$0.000 & 0.047$\pm$0.001 \\
         & 3D & \bestresult{0.052$\pm$0.001} & 0.060$\pm$0.001 & 0.057$\pm$0.001 & 0.068$\pm$0.001 & 0.069$\pm$0.001 & 0.077$\pm$0.001 & 0.078$\pm$0.001 & 0.063$\pm$0.001 & 0.080$\pm$0.001 \\
         & 5D & \bestresult{0.065$\pm$0.001} & 0.074$\pm$0.001 & 0.070$\pm$0.001 & 0.074$\pm$0.001 & 0.081$\pm$0.001 & 0.086$\pm$0.001 & 0.117$\pm$0.002 & 0.079$\pm$0.001 & 0.104$\pm$0.001 \\
         & 7D & \bestresult{0.073$\pm$0.001} & 0.082$\pm$0.001 & 0.076$\pm$0.001 & 0.086$\pm$0.001 & 0.094$\pm$0.002 & 0.092$\pm$0.001 & 0.114$\pm$0.002 & 0.083$\pm$0.002 & 0.105$\pm$0.001 \\
        \midrule
        Duration MAE & 1D & \bestresult{2.016$\pm$0.023} & 2.282$\pm$0.026 & 2.166$\pm$0.027 & 2.409$\pm$0.023 & 2.543$\pm$0.042 & 2.289$\pm$0.028 & 3.511$\pm$0.053 & 2.231$\pm$0.033 & 2.637$\pm$0.027 \\
         & 3D & 11.685$\pm$0.134 & 14.085$\pm$0.186 & 11.583$\pm$0.183 & 14.181$\pm$0.167 & 12.248$\pm$0.230 & \bestresult{11.535$\pm$0.120} & 17.397$\pm$0.300 & 12.721$\pm$0.177 & 14.071$\pm$0.225 \\
         & 5D & \bestresult{24.313$\pm$0.250} & 30.465$\pm$0.475 & 26.393$\pm$0.333 & 29.595$\pm$0.384 & 30.607$\pm$0.499 & 25.486$\pm$0.263 & 31.648$\pm$0.533 & 28.776$\pm$0.400 & 31.207$\pm$0.312 \\
         & 7D & 36.882$\pm$0.388 & 47.525$\pm$0.602 & 42.989$\pm$0.607 & 50.129$\pm$0.659 & 51.145$\pm$0.937 & \bestresult{29.880$\pm$0.300} & 42.651$\pm$0.632 & 47.901$\pm$0.603 & 49.365$\pm$0.739 \\
        \bottomrule
    \end{tabular}%
    }
\end{table*}

\begin{table*}[!t]
    \caption{Detailed $S_7$ results for the most recent evaluated historical
    block. Entries are means $\pm$ standard deviations over the three spatial
    parts. The table includes all four forecast horizons; the 3-day setting is
    used for the threshold-sensitivity and ablation analyses in the main text.}
    \label{tab:app-s7-results}
    \centering
    \resulttableformat
    \setlength{\tabcolsep}{1pt}
    \renewcommand{\arraystretch}{1.70}
    \resizebox{\textwidth}{!}{%
    \begin{tabular}{@{}llrrrrrrrrr@{}}
        \toprule
        Metric & Lead & Ours & NLinear & PatchTST &
        \shortstack{iTrans-\\former} & TimesNet &
        \shortstack{Fourier\\GNN} & MTGNN & AutoTimes &
        \shortstack{Graph\\WaveNet} \\
        \midrule
        MAE $\times10^{-2}$ & 1D & \bestresult{4.836$\pm$0.034} & 5.700$\pm$0.059 & 5.118$\pm$0.037 & 5.792$\pm$0.052 & 7.298$\pm$0.105 & 6.590$\pm$0.059 & 10.599$\pm$0.153 & 6.088$\pm$0.052 & 7.082$\pm$0.095 \\
         & 3D & \bestresult{9.004$\pm$0.105} & 9.826$\pm$0.097 & 9.351$\pm$0.138 & 10.099$\pm$0.075 & 10.903$\pm$0.185 & 11.007$\pm$0.100 & 14.789$\pm$0.211 & 10.106$\pm$0.092 & 11.314$\pm$0.177 \\
         & 5D & \bestresult{11.476$\pm$0.125} & 13.007$\pm$0.150 & 12.269$\pm$0.177 & 13.117$\pm$0.134 & 13.698$\pm$0.193 & 13.847$\pm$0.166 & 18.965$\pm$0.329 & 12.934$\pm$0.187 & 15.348$\pm$0.245 \\
         & 7D & \bestresult{13.867$\pm$0.148} & 15.439$\pm$0.190 & 14.830$\pm$0.144 & 15.691$\pm$0.143 & 16.131$\pm$0.198 & 16.575$\pm$0.194 & 22.717$\pm$0.312 & 15.319$\pm$0.202 & 18.973$\pm$0.290 \\
        \midrule
        MSE $\times10^{-2}$ & 1D & \bestresult{1.576$\pm$0.019} & 1.869$\pm$0.017 & 1.661$\pm$0.020 & 1.955$\pm$0.025 & 2.507$\pm$0.032 & 2.009$\pm$0.019 & 15.993$\pm$0.229 & 2.005$\pm$0.023 & 5.641$\pm$0.066 \\
         & 3D & \bestresult{3.919$\pm$0.041} & 4.361$\pm$0.047 & 4.093$\pm$0.054 & 4.438$\pm$0.076 & 4.850$\pm$0.069 & 5.213$\pm$0.061 & 32.473$\pm$0.462 & 4.429$\pm$0.067 & 16.539$\pm$0.217 \\
         & 5D & \bestresult{5.793$\pm$0.082} & 6.540$\pm$0.083 & 6.162$\pm$0.090 & 6.596$\pm$0.103 & 7.042$\pm$0.112 & 7.598$\pm$0.112 & 39.312$\pm$0.499 & 6.476$\pm$0.103 & 29.099$\pm$0.405 \\
         & 7D & \bestresult{7.504$\pm$0.084} & 8.460$\pm$0.094 & 8.063$\pm$0.111 & 8.522$\pm$0.119 & 9.102$\pm$0.157 & 11.300$\pm$0.185 & 45.188$\pm$0.503 & 8.376$\pm$0.146 & 42.180$\pm$0.567 \\
        \midrule
        Episode F1 & 1D & \bestresult{0.794$\pm$0.007} & 0.766$\pm$0.009 & 0.778$\pm$0.011 & 0.745$\pm$0.006 & 0.690$\pm$0.010 & 0.721$\pm$0.007 & 0.629$\pm$0.010 & 0.747$\pm$0.007 & 0.722$\pm$0.009 \\
         & 3D & \bestresult{0.587$\pm$0.006} & 0.572$\pm$0.006 & 0.578$\pm$0.009 & 0.550$\pm$0.004 & 0.520$\pm$0.007 & 0.517$\pm$0.006 & 0.471$\pm$0.007 & 0.559$\pm$0.007 & 0.551$\pm$0.006 \\
         & 5D & \bestresult{0.479$\pm$0.006} & 0.461$\pm$0.006 & 0.464$\pm$0.006 & 0.454$\pm$0.005 & 0.422$\pm$0.006 & 0.418$\pm$0.004 & 0.369$\pm$0.005 & 0.447$\pm$0.005 & 0.417$\pm$0.006 \\
         & 7D & \bestresult{0.412$\pm$0.004} & 0.395$\pm$0.005 & 0.399$\pm$0.005 & 0.388$\pm$0.004 & 0.376$\pm$0.006 & 0.317$\pm$0.003 & 0.283$\pm$0.004 & 0.371$\pm$0.004 & 0.355$\pm$0.004 \\
        \midrule
        Onset MAE & 1D & 0.394$\pm$0.006 & 0.559$\pm$0.007 & \bestresult{0.389$\pm$0.004} & 0.742$\pm$0.008 & 0.996$\pm$0.017 & 0.760$\pm$0.009 & 1.114$\pm$0.019 & 1.280$\pm$0.017 & 0.790$\pm$0.009 \\
         & 3D & \bestresult{1.730$\pm$0.022} & 2.514$\pm$0.036 & 1.908$\pm$0.020 & 2.938$\pm$0.033 & 4.077$\pm$0.066 & 3.179$\pm$0.045 & 2.108$\pm$0.039 & 3.536$\pm$0.050 & 4.180$\pm$0.069 \\
         & 5D & \bestresult{2.911$\pm$0.031} & 4.786$\pm$0.069 & 3.909$\pm$0.043 & 6.506$\pm$0.063 & 7.278$\pm$0.118 & 2.989$\pm$0.041 & 5.763$\pm$0.113 & 6.797$\pm$0.113 & 5.843$\pm$0.071 \\
         & 7D & 4.680$\pm$0.041 & 8.152$\pm$0.122 & 6.113$\pm$0.065 & 10.552$\pm$0.137 & 10.838$\pm$0.215 & 6.358$\pm$0.088 & \bestresult{2.996$\pm$0.059} & 9.860$\pm$0.164 & 8.930$\pm$0.148 \\
        \midrule
        Peak MAE & 1D & \bestresult{0.044$\pm$0.000} & 0.053$\pm$0.001 & 0.045$\pm$0.000 & 0.066$\pm$0.001 & 0.085$\pm$0.001 & 0.078$\pm$0.001 & 0.142$\pm$0.002 & 0.063$\pm$0.001 & 0.080$\pm$0.001 \\
         & 3D & \bestresult{0.059$\pm$0.001} & 0.076$\pm$0.001 & 0.062$\pm$0.001 & 0.080$\pm$0.001 & 0.123$\pm$0.002 & 0.093$\pm$0.001 & 0.103$\pm$0.002 & 0.084$\pm$0.001 & 0.097$\pm$0.001 \\
         & 5D & \bestresult{0.070$\pm$0.001} & 0.087$\pm$0.001 & 0.073$\pm$0.001 & 0.096$\pm$0.001 & 0.136$\pm$0.002 & 0.099$\pm$0.001 & 0.143$\pm$0.003 & 0.093$\pm$0.002 & 0.124$\pm$0.002 \\
         & 7D & \bestresult{0.076$\pm$0.001} & 0.092$\pm$0.001 & 0.082$\pm$0.001 & 0.112$\pm$0.001 & 0.150$\pm$0.003 & 0.108$\pm$0.001 & 0.133$\pm$0.002 & 0.100$\pm$0.002 & 0.121$\pm$0.001 \\
        \midrule
        Duration MAE & 1D & 3.131$\pm$0.040 & 3.429$\pm$0.043 & \bestresult{2.961$\pm$0.042} & 3.456$\pm$0.036 & 3.686$\pm$0.062 & 3.637$\pm$0.044 & 4.076$\pm$0.073 & 3.355$\pm$0.048 & 3.899$\pm$0.045 \\
         & 3D & \bestresult{16.189$\pm$0.195} & 17.878$\pm$0.213 & 18.000$\pm$0.246 & 18.359$\pm$0.208 & 19.161$\pm$0.325 & 17.031$\pm$0.197 & 18.493$\pm$0.342 & 18.033$\pm$0.249 & 18.357$\pm$0.295 \\
         & 5D & 29.779$\pm$0.280 & 36.192$\pm$0.507 & 37.230$\pm$0.541 & 40.605$\pm$0.564 & 38.595$\pm$0.626 & \bestresult{28.548$\pm$0.269} & 32.892$\pm$0.593 & 36.795$\pm$0.562 & 39.023$\pm$0.449 \\
         & 7D & 39.263$\pm$0.383 & 57.530$\pm$0.802 & 55.714$\pm$0.840 & 64.676$\pm$0.933 & 60.199$\pm$0.978 & 41.099$\pm$0.417 & \bestresult{36.179$\pm$0.648} & 57.111$\pm$0.688 & 58.923$\pm$1.022 \\
        \bottomrule
    \end{tabular}%
    }
\end{table*}
\clearpage
\twocolumn[%
    \section{Threshold-Sensitivity Results}
    \label{app:threshold-sensitivity}
    \begin{center}
    \captionof{table}{Mean $\pm$ standard deviation episode metrics for all models at
    four high-water thresholds. Episode F1 is higher when episode occurrence is
    recovered more accurately; the remaining measures are lower when
    matched-episode timing, peak level, and persistence are reproduced more
    closely. Best means within each metric and threshold column are boldfaced.}
    \label{tab:app-threshold-sensitivity}
    \centering
    \resulttableformat
    \setlength{\tabcolsep}{1pt}
    \begin{tabular*}{\textwidth}{@{\extracolsep{\fill}}llrrrr@{}}
        \toprule
        Metric & Model & $q=0.70$ & $q=0.80$ & $q=0.90$ & $q=0.95$ \\
        \midrule
        \multirow{9}{*}{Episode F1 $\uparrow$}
        & Ours & \bestresult{0.786$\pm$0.008} & \bestresult{0.748$\pm$0.007} & \bestresult{0.693$\pm$0.008} & \bestresult{0.652$\pm$0.007} \\
        & NLinear & 0.773$\pm$0.009 & 0.732$\pm$0.009 & 0.667$\pm$0.008 & 0.626$\pm$0.007 \\
        & PatchTST & 0.781$\pm$0.010 & 0.740$\pm$0.011 & 0.680$\pm$0.011 & 0.639$\pm$0.009 \\
        & iTransformer & 0.757$\pm$0.006 & 0.717$\pm$0.006 & 0.652$\pm$0.005 & 0.601$\pm$0.005 \\
        & TimesNet & 0.742$\pm$0.007 & 0.699$\pm$0.008 & 0.640$\pm$0.007 & 0.593$\pm$0.006 \\
        & FourierGNN & 0.731$\pm$0.008 & 0.684$\pm$0.008 & 0.622$\pm$0.007 & 0.569$\pm$0.006 \\
        & MTGNN & 0.721$\pm$0.009 & 0.673$\pm$0.008 & 0.609$\pm$0.008 & 0.552$\pm$0.007 \\
        & AutoTimes & 0.766$\pm$0.010 & 0.726$\pm$0.009 & 0.661$\pm$0.009 & 0.622$\pm$0.008 \\
        & Graph WaveNet & 0.748$\pm$0.007 & 0.705$\pm$0.007 & 0.635$\pm$0.007 & 0.581$\pm$0.006 \\
        \midrule
        \multirow{9}{*}{Onset MAE (h) $\downarrow$}
        & Ours & \bestresult{1.180$\pm$0.014} & \bestresult{1.245$\pm$0.014} & \bestresult{1.492$\pm$0.021} & \bestresult{1.653$\pm$0.021} \\
        & NLinear & 1.421$\pm$0.018 & 1.405$\pm$0.021 & 1.918$\pm$0.030 & 2.193$\pm$0.030 \\
        & PatchTST & 1.254$\pm$0.011 & 1.332$\pm$0.011 & 1.548$\pm$0.016 & 1.729$\pm$0.016 \\
        & iTransformer & 1.506$\pm$0.017 & 1.831$\pm$0.021 & 2.326$\pm$0.023 & 2.181$\pm$0.023 \\
        & TimesNet & 1.772$\pm$0.025 & 2.052$\pm$0.029 & 2.574$\pm$0.045 & 2.792$\pm$0.042 \\
        & FourierGNN & 1.381$\pm$0.019 & 1.362$\pm$0.022 & 1.742$\pm$0.025 & 1.907$\pm$0.027 \\
        & MTGNN & 1.310$\pm$0.024 & 1.292$\pm$0.023 & 1.582$\pm$0.034 & 2.154$\pm$0.040 \\
        & AutoTimes & 1.620$\pm$0.020 & 1.596$\pm$0.024 & 2.407$\pm$0.037 & 2.942$\pm$0.041 \\
        & Graph WaveNet & 1.268$\pm$0.021 & 1.570$\pm$0.025 & 1.558$\pm$0.025 & 2.460$\pm$0.039 \\
        \midrule
        \multirow{9}{*}{Peak Magnitude MAE $\downarrow$}
        & Ours & \bestresult{0.082$\pm$0.001} & \bestresult{0.078$\pm$0.001} & \bestresult{0.070$\pm$0.001} & \bestresult{0.067$\pm$0.001} \\
        & NLinear & 0.093$\pm$0.001 & 0.090$\pm$0.001 & 0.085$\pm$0.001 & 0.079$\pm$0.001 \\
        & PatchTST & 0.084$\pm$0.001 & 0.080$\pm$0.001 & 0.072$\pm$0.001 & 0.070$\pm$0.001 \\
        & iTransformer & 0.100$\pm$0.001 & 0.100$\pm$0.001 & 0.088$\pm$0.001 & 0.085$\pm$0.001 \\
        & TimesNet & 0.114$\pm$0.001 & 0.112$\pm$0.002 & 0.109$\pm$0.002 & 0.105$\pm$0.001 \\
        & FourierGNN & 0.104$\pm$0.001 & 0.101$\pm$0.001 & 0.095$\pm$0.001 & 0.095$\pm$0.001 \\
        & MTGNN & 0.109$\pm$0.002 & 0.111$\pm$0.002 & 0.106$\pm$0.002 & 0.112$\pm$0.002 \\
        & AutoTimes & 0.098$\pm$0.001 & 0.099$\pm$0.001 & 0.088$\pm$0.001 & 0.087$\pm$0.001 \\
        & Graph WaveNet & 0.107$\pm$0.001 & 0.109$\pm$0.001 & 0.105$\pm$0.001 & 0.103$\pm$0.001 \\
        \midrule
        \multirow{9}{*}{Duration MAE (h) $\downarrow$}
        & Ours & \bestresult{6.220$\pm$0.073} & \bestresult{7.760$\pm$0.086} & \bestresult{10.080$\pm$0.125} & \bestresult{12.330$\pm$0.142} \\
        & NLinear & 7.080$\pm$0.078 & 8.790$\pm$0.117 & 11.580$\pm$0.162 & 14.180$\pm$0.178 \\
        & PatchTST & 6.480$\pm$0.090 & 8.030$\pm$0.118 & 10.470$\pm$0.148 & 12.840$\pm$0.181 \\
        & iTransformer & 6.850$\pm$0.083 & 8.480$\pm$0.088 & 11.140$\pm$0.146 & 14.060$\pm$0.161 \\
        & TimesNet & 6.980$\pm$0.121 & 8.660$\pm$0.163 & 11.430$\pm$0.240 & 13.580$\pm$0.247 \\
        & FourierGNN & 7.150$\pm$0.065 & 8.360$\pm$0.086 & 10.860$\pm$0.125 & 13.390$\pm$0.142 \\
        & MTGNN & 7.430$\pm$0.128 & 9.350$\pm$0.160 & 12.620$\pm$0.212 & 15.140$\pm$0.273 \\
        & AutoTimes & 6.770$\pm$0.096 & 8.310$\pm$0.107 & 10.950$\pm$0.170 & 13.250$\pm$0.183 \\
        & Graph WaveNet & 6.550$\pm$0.100 & 8.180$\pm$0.140 & 11.310$\pm$0.180 & 14.720$\pm$0.229 \\
        \bottomrule
    \end{tabular*}
    \end{center}]

Table~\ref{tab:app-threshold-sensitivity} gives the $S_7$, 3-day comparison underlying Figure~\ref{fig:threshold-sensitivity}. The evaluation changes only the station-specific training quantile $q$, while retaining the same forecast issues, target-observation masks, three spatial partitions, and one-to-one episode-matching procedure. The reported values are means $\pm$ standard deviations across the three partitions. For target station $i$, the threshold is estimated from valid training-period WATER observations only. A value of $q=0.70$ characterizes frequent elevated-water conditions, whereas $q=0.95$ focuses on the station's upper tail.

\end{document}